\documentclass[sigconf]{acmart}
\usepackage[dvipsnames]{xcolor}
\usepackage{stmaryrd}
\usepackage{amsmath,amsfonts,amsbsy}
\usepackage[mathscr]{euscript}
\usepackage{hyperref}
\usepackage{url}
\usepackage{booktabs}
\usepackage{multirow}
\usepackage{float}
\usepackage[ruled,vlined,linesnumbered]{algorithm2e}
\usepackage{graphicx}
\usepackage{subfigure}
\usepackage{balance}
\usepackage{graphicx}
\usepackage{booktabs}
\usepackage{makecell}
\usepackage{array}
\usepackage{enumitem}
\setlist[itemize]{noitemsep, topsep=3pt}

\usepackage[capitalise,nameinlink]{cleveref}
\crefname{section}{Sec.}{Sec.}
\Crefname{section}{Section}{Sections}
\crefname{listing}{List.}{List.}
\crefname{listing}{Listing}{Listings}
\Crefname{listing}{Listing}{Listings}
\crefname{lstlisting}{Listing}{Listings}
\Crefname{lstlisting}{Listing}{Listings}

\newcommand{\codeurl}{\url{https://bit.ly/3NWr6dd}}

\newcommand{\hide}[1]{}

\newcommand{\T}[1]{\boldsymbol{\mathscr{#1}}}

\newcommand{\mat}[1]{\mathbf{#1}}
\newcommand{\vect}[1]{\mathbf{#1}}

\AtBeginDocument{%
  }

\setcopyright{acmlicensed}
\copyrightyear{2026}
\acmYear{2026}

\acmConference[Preprint]{February 12}{2026}

\begin{document}

\title{Transforming Behavioral Neuroscience Discovery with In-Context Learning and AI-Enhanced Tensor Methods}

\settopmatter{authorsperrow=4}
\author{Paimon Goulart}
\authornote{All three authors contributed equally to this research.}
\affiliation{%
    \institution{UC Riverside}
  \institution{Computer Science \& Engineering}
  \city{Riverside, CA}
  \country{USA}
}
\email{pgoul002@ucr.edu}

\author{Jordan Steinhauser}
\authornotemark[1]
\affiliation{%
  \institution{UC Riverside}
  \institution{Psychology}
  \city{Riverside, CA}
  \country{USA}
}
\email{jstei007@ucr.edu}

\author{Dawon Ahn}
\authornotemark[1]
\affiliation{%
  \institution{UC Riverside}
  \institution{Computer Science \& Engineering}
  \city{Riverside, CA}
  \country{USA}
}
\email{dahn017@ucr.edu}

\author{Kylene Shuler}
\affiliation{%
  \institution{UC Riverside}
  \institution{Psychology}
  \city{Riverside, CA}
  \country{USA}
}
\email{kshul004@ucr.edu}

\author{Edward Korzus}
\affiliation{%
  \institution{UC Riverside}
  \institution{Psychology}
  \city{Riverside, CA}
  \country{USA}
}
\email{edward.korzus@ucr.edu}

\author{Jia Chen}
\affiliation{%
  \institution{UC Riverside}
  \institution{Electrical \& Computer Engineering}
  \city{Riverside, CA}
  \country{USA}
}
\email{jiac@ucr.edu}

\author{Evangelos~E.\,Papalexakis}
\affiliation{%
  \institution{UC Riverside}
  \institution{Computer Science \& Engineering}
  \city{Riverside, CA}
  \country{USA}
}
\email{epapalex@cs.ucr.edu}

\renewcommand{\shortauthors}{Goulart et al.}

\begin{abstract}
Scientific discovery pipelines typically involve complex, rigid, and time-consuming processes, from data preparation to analyzing and interpreting findings. Recent advances in AI have the potential to transform such pipelines in a way that domain experts can focus on interpreting and understanding findings, rather than debugging rigid pipelines or manually annotating data.

In this paper, as part of an active collaboration between data science/AI researchers and behavioral neuroscientists, we showcase an example AI-enhanced pipeline, specifically designed to transform and accelerate the way that the domain experts in the team are able to gain insights out of experimental data. The application at hand is in the domain of behavioral neuroscience, studying fear generalization in mice, an important problem whose progress can advance our understanding of clinically significant and often debilitating conditions such as PTSD (Post-Traumatic Stress Disorder). 

From a technical point of view, we identify the emerging paradigm of "In-Context Learning" (ICL) as a suitable interface for domain experts to automate parts of their pipeline without the need for or familiarity with AI model training and fine-tuning, and showcase its remarkable efficacy in data preparation and pattern interpretation. At the same time, we introduce novel AI-enhancements to tensor decomposition, a class of methods that has been shown to be effective in behavioral neuroscience alongside a vast array of other domains, which allow for more seamless pattern discovery from the heterogeneous data in our application.

We thoroughly evaluate every component of our proposed pipeline experimentally, showcasing its competitive or superior performance compared to what is standard practice in the domain, as well as against reasonable ML baselines that do not fall under the ICL paradigm, to ensure that we are not compromising performance in our quest for a seamless and easy-to-use interface for domain experts. Finally, we demonstrate effective discovery, with results validated by the domain experts in the team.
\end{abstract}


\maketitle

\section{Introduction}

Behavioral Neuroscience is a prime example of a discipline whose potential can be maximized with the use of data-driven approaches and AI methods. The specific focus of this work is understanding fear discrimination, and more specifically, how different mouse subjects learn to generalize fear across environments, and what factors may lead to fear over-generalization, a condition whose mechanism can help shed light to often debilitating conditions in humans such as Post-Traumatic Stress Disorder (PTSD).
The use of mouse models proves to be valuable when investigating fear due to the high preservation of neuronal circuitry involved in fear between mice and humans \cite{tovote2015neuronal}, and also allows researchers to conduct thorough investigations into threat processing. In this problem, we are interested in both neural patterns (measured using calcium imaging \cite{ghosh2011miniaturized}) as well as behavioral patterns (captured through videos and typically manually encoded by experts). 

This work is a product of an interdisciplinary research team of computer scientists with expertise in AI and data science methods and behavioral neuroscientists with extensive domain expertise and working knowledge of AI and data science methods from a user's perspective. We present an end-to-end use case of how AI methods can transform a typical behavioral neuroscience research pipeline, followed not only by the domain experts in our team but also widely accepted and followed by researchers in the domain \cite{cai2016shared, patel2022longitudinal, jacob2018compact, besnard2019dorsolateral, marin2024sex, zhou2018efficient, agetsuma2023activity, lecca2020heterogeneous, botta2020amygdala}. 

Introducing data science methods at-large in behavioral neuroscience applications is not a new proposition. However, the most impactful such attempts focus on a very specific part of the research pipeline that relates to the data analysis. For example, there have been successful attempts in going beyond ``standard'' statistical significance tests \cite{deadwyler1997significance, krimer2005cluster} and introducing knowledge discovery methods in analyzing brain measurements using tensor methods \cite{williams2018unsupervised, faghiri2024analysis} and graphs \cite{fornito2016fundamentals, garcia2018applications} among others. Such attempts are extremely valuable in that they allow for the discovery of potentially previously-unknown patterns, which may lead the domain experts to produce novel hypotheses. 

However, the impact of such attempts is still very narrow and not transformative. In order for such complex knowledge discovery methods to have a chance to produce meaningful results, there needs to be an ``analysis-ready'' or ``AI-ready'' \cite{carter2023advanced} dataset. At the same time, in order for a domain expert to be able to make use of the discovered patterns, they need to have working knowledge and understanding of the knowledge discovery method or the AI model that is being used for the analysis. Those two desired data entail a long, tedious, and time-consuming data preparation process while requiring significant AI knowledge on the part of the domain expert. As a result, such attempts are not scalable to the greater population of domain experts, however rich their potential may be.

In this paper, we identify different parts of the behavioral neuroscience pipeline that can be transformed with AI and knowledge discovery methods and we introduce an end-to-end pipeline that does so in a way that it leverages the domain expert's input and knowledge when necessary while decoupling their involvement from processes that are either manual, time-consuming, and tedious or require significant computing knowledge (or both) so that they can focus on understanding, evaluating, and acting upon the results of the pipeline.
We introduce the following components of the pipeline, aligned with our technical contributions:
\begin{itemize}
\item {\bf In-Context Data Preparation}: We identify In-Context Learning (ICL) as a great candidate for allowing domain experts to interface with cutting-edge AI foundation models without the need for advanced knowledge in model training. We demonstrate how tasks in data preparation can be automated as such, and we introduce the novel concept of {\it Autoregressive In-Context Learning (AR-ICL)}.
\item {\bf Neural Tensor Analysis}:
We enhance tensor analysis by leveraging advanced tensor decomposition models 
for discovering hidden patterns in the coupled data with multiple shared attributes, 
and identifying the most contributing patterns across different data sources.
\item {\bf AI-driven Pattern Interpretation}: We leverage domain knowledge, Retrieval-Augmented Generation (RAG), and In-Context Learning in understanding and interpreting the discovered patterns.
\end{itemize}

From the application's point of view, we conduct an end-to-end evaluation of our proposed pipeline and identify patterns in the data that have the potential to drive discovery in the problem of fear (over)generalization and discrimination.
We make our {\bf source code} and {\bf datasets}  publicly available at \codeurl.

%
\section{Traditional Worfklow}
Prior work in the field of behavioral neuroscience that investigates neuronal network activity, has relied on available open-source scripts that address single parts of the complex data analysis process to be strung together in a coherent pipeline. Although this has worked for research teams in the past, this method requires programming knowledge that many domain experts do not possess. Additionally, this standard way of processing datasets has proven rigid, as off-the-shelf programs for brain data analysis still need to be updated and adjusted according to the precise data representation generated by the experimental design, and time consuming, as these updates to parameters and learning to adjust how data is input and the desired results output can be challenging and will take time for researchers with minimal programming capabilities. Improvements to the current standard practice, such as the ones proposed in this paper, will provide domain experts an efficient and effective pipeline to process their data, run analyses, and interpret results, while  ultimately allowing more focus by the researchers to be placed on discovery questions and big-picture thinking.

\noindent{\bf Data Collection} Neuronal activity of mice was recorded using one-photon calcium imaging \cite{grienberger2012imaging}. Two-month old mice underwent a series of surgical procedures including a viral injection into the right prelimbic region (PL) of the medial prefrontal cortex to express a genetically encoded calcium indicator \cite{chen2013ultrasensitive}, which fluoresces upon calcium binding. Then a 1mm prism lens was chronically implanted lateral to the viral injection site. During recording sessions, a head-mounted miniaturized microscope was transiently attached to an aluminum docking plate, enabling the visualization of neuronal activity in PL through changes in fluorescence \cite{zhou2023deep, stringer2019computational}. Section \ref{sec:discovery} describes specifics for our fear conditioning and discrimination application.
Overall, this set-up is considered standard in the domain. 

{\noindent{\bf Data Preparation}
For the most part, data preparation is either done manually or via scripts that require significant computing familiarity. In our specific case, labeling of mouse behavior is done by visual inspection of the videos by a domain expert, while conversion of calcium imaging to different types of workable inputs to data analysis tools is done by pipelines such as CNMF \cite{pnevmatikakis2016simultaneous} and a cell registration process \cite{sheintuch2017tracking}.

\noindent{\bf Data Analysis}
Early experiments performed cluster analysis to divide neurons, whose activity was recorded using eletrophysiology, into discrete categories of regular-spiking and fast-spiking neurons, and ANOVA to determine between-group similarities and differences \cite{krimer2005cluster}. Later, researchers worked to use community detection algorithms to generate graphs of interconnected units within the brain \cite{garcia2018applications}. Additionally, Williams and colleagues novelly applied tensor component analysis (TCA) to neuronal activity data collected from freely behaving animals using NeuroPixels \cite{williams2018unsupervised}. This application was revolutionary for the field as it allowed neuroscience researchers to discover latent patterns in neuronal activity data in an unsupervised manner; however, it did not address important aspects of using TCA on data collected during learning and cognition and did not clearly identify reliable methods to determining the number of components.

Prior work involving a subset of our team independently utilized TCA and community analysis to discover groups of neurons that are co-engaged during unique environments and epochs of a fear conditioning and discrimination paradigm, in order to discover populations of neurons that are implicated in threat detection and safety learning \cite{pastore2024prefrontal}. This and previous work have advanced computational techniques for behavioral neuroscience, but their accurate application often requires highly interdisciplinary expertise. Streamlined pipelines, such as the one proposed here, can reduce this complexity and enable wider adoption across diverse areas.

\noindent{\bf Result Analysis}
Finally, the analysis of results is done purely manually and depending on the analysis tools used it requires substantial understanding of those tools by the domain expert.

\hide
{
\noindent{\bf Data Preparation and Result Analysis}
For the most part, data preparation and analysis is either done manually or via scripts that require significant computing familiarity. In our specific case, labeling of mouse behavior is done by visual inspection of the videos by a domain expert, while conversion of calcium imaging to different types of workable inputs to data analysis tools is done by pipelines such as CNMF \cite{pnevmatikakis2016simultaneous} and a cell registration process \cite{sheintuch2017tracking}. Finally, the analysis of results is done purely manually and depending on the analysis tools used it requires substantial understanding of those tools by the domain expert. 
}

\begin{figure*}[ht!]
    \begin{center}
        \includegraphics[width =.9\textwidth]{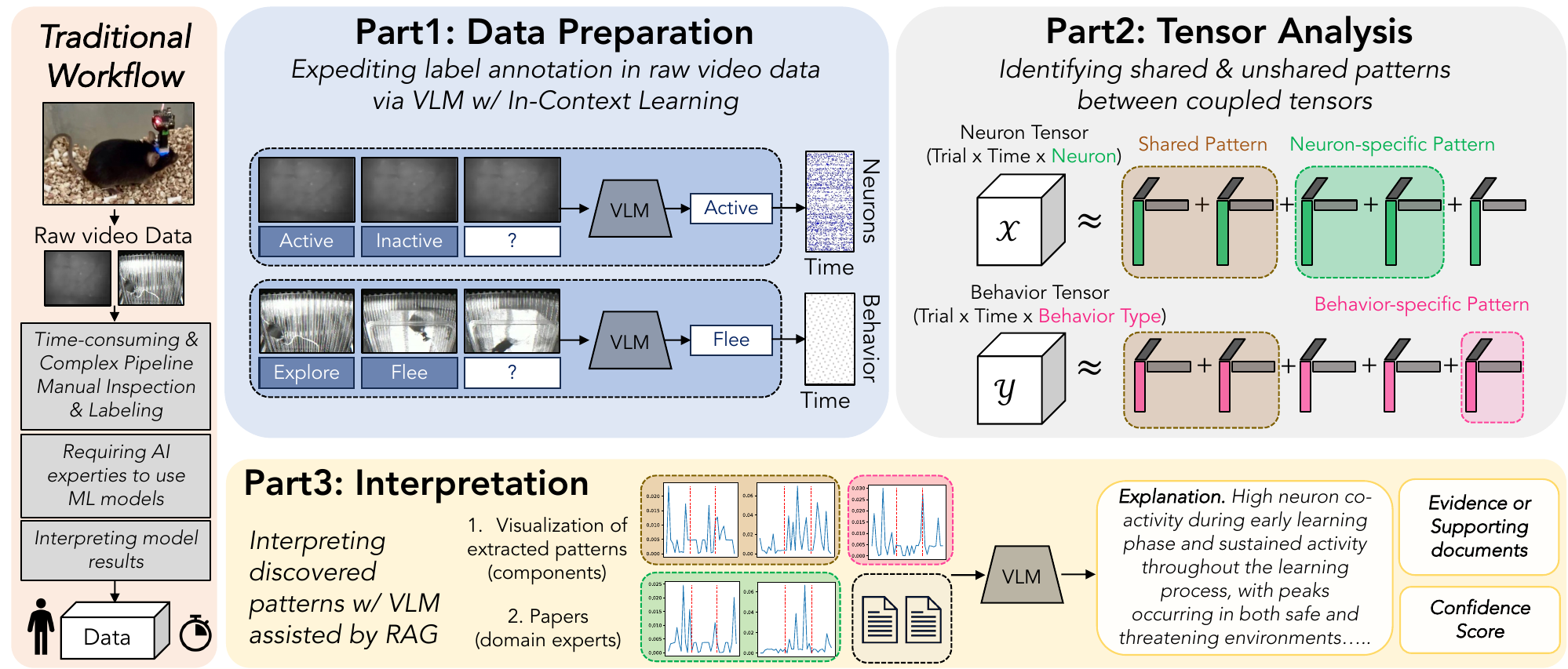}
        \caption{Overview of the proposed pipeline. 
        Our pipeline streamlines time-consuming and expertise-intensive steps, easing the workload for scientists,
        in contrast to conventional workflow.
            }
        \label{fig:overview}
    \end{center}
\end{figure*}
\section{Proposed Method} 
We propose an AI-enhanced pipeline that automates time-consuming and expertise-intensive steps in behavioral neuroscience workflows.
As illustrated in \Cref{fig:overview}, the pipeline consists of three main parts:
(1) In-context data preparation, (2) AI-enhanced Tensor Analysis, and
(3) AI-driven pattern interpretation.
In the first part, the pipeline provides a user-friendly interface that enables scientists to generate clean datasets from raw video data without requiring knowledge of AI models.
By leveraging In-Context Learning (ICL), it removes the need for manual inspection or labeling.
The second part uses a neural tensor decomposition method to enhance analysis of while maintaining interpretability of classical tensor models.
In the third part, an agent facilitates interpretation of results from tensor analysis by
leveraging VLM and retrieval-augmented generation (RAG). Details for each part follow.

\subsection{In-Context Data Preparation} \label{step1:icl}
In-Context Learning (ICL) is an emerging technique that allows LLMs and VLMs to perform domain specific classification, regression, structured prediction, and other tasks without the need for fine-tuning or domain specific training \cite{NEURIPS2020_1457c0d6, min2022rethinkingroledemonstrationsmakes, goulart2024can, ferber2024incontextlearningenablesmultimodal, kim2024videoiclconfidencebasediterativeincontext, zhang2025videoincontextlearningautoregressive}. This is done by providing an LLM or VLM a handful of different examples of input output pairs $S = \{(x_1, y_1), \dots, (x_k, y_k)\}$ for a specific task, and then asking it to predict the next unseen target from a given input $x_{k+1}$ where $k$ represents the number of examples.

ICL requires minimal knowledge of how LLMs or VLMs operate internally, in contrast to fine-tuning, which requires technical expertise, significant computational resources, and large amounts of labeled training data. ICL is also substantially easier to set up than other computer vision pipelines based on models such as DINO-V2, which often require specialized training procedures and may again also need fine-tuning for the specific task. Additionally, prior work has shown that ICL can achieve competitive, and in some cases state-of-the-art, performance across vision-language tasks \cite{goulart2025preliminaryusevisionlanguage}. This allows domain experts from non machine learning related fields to utilize this tool with ease.

\subsubsection{Behavioral Video Labeling}

Our first task focuses on automated labeling of mouse behavioral videos. Following prior work \cite{goulart2025preliminaryusevisionlanguage}, we consider second by second classification of behavior from mice experiment recordings. Each second of the video is given one of the following behavior labels: freezing, fleeing, or grooming/exploring. Manually producing these annotations can take an extremely long time, requiring researchers or volunteers to spend a significant amount of resources in order to review and label a single video. This manual labor limits the scale of these behavioral studies and creates a demand for accurate automated labeling methods. Previous work has shown that a VLM provided with in-context learning (ICL) examples, along with frame splitting to avoid VLM frame sampling which can cause crucial frames to be missed, can achieve competitive performance on this task, surpassing baselines such as nearest neighbor DINO-V2 \cite{imam2025multimodalllmsvisualtemporal, caron2021emergingpropertiesselfsupervisedvision}.

However, we observed two key limitations in this pipeline. First, behaviors often continue across different seconds: a mouse may continue the same behavior from the previous second, and behaviors can also begin in one second and leak into the next. Treating each second independently ignores this continuity. Second, treating each second independently can cause the VLM to output drastic labeling inconsistencies over a short period of time. For example, normal exploring movement that briefly speeds up may be mislabeled as fleeing, even though the overall behavior has not changed. Such temporal inconsistency can reduce the reliability and scientific interpretability of model outputs.

One might think that this issue can be resolved by providing the VLM with longer clips or multiple seconds at once, however in practice this approach has been found to be extremely unreliable. This is also found to be the case in prior work where VLMs have been observed to struggle with detailed temporal reasoning in videos \cite{imam2025multimodalllmsvisualtemporal}. For example, behaviors might be correctly identified, but attributed to the wrong second, or the model would just label the same behavior across all seconds.

To address these issues, we introduce \textbf{Autoregressive in-context learning (AR-ICL)}. In AR-ICL, the model is not only provided with the original ICL demonstration examples, but also with its own most recent prediction 
as additional in-context information. 
When predicting the label for the current 1-second chunk $x^{(t)}$, the prompt, denoted by $S_{AR}$, includes the previous second chuck $x^{(t-1)}$ and its predicted label $\hat{y}^{(t-1)}$ along with the fixed examples $\{(x_i, y_i)\}_{i=1}^k$ where $k$ represents the number of fixed ICL examples, i.e.,
$S_{AR} = \{(x_1, y_1), \dots, (x_k, y_k),(x^{(t-1)}, \hat{y}^{(t-1)})\}$. This allows the VLM to reason about how the current behavior relates to what was happening immediately before. Intuitively, this encourages temporal smoothness and reduces abrupt label changes that contradict what just previously happened. Although AR-ICL is specifically applied towards the task of behavioral labeling here, we believe this strategy can be broadly applicable to sequential prediction tasks where outputs are temporally dependent (e.g. other video prediction tasks and time-series classification).

In addition to AR-ICL, we also provide the next one-second video segment \(x^{(t+1)}\) as unlabeled visual context. This next second context can help with transitional behaviors that can go on for multiple seconds, while keeping the autoregressive structure centered on past predictions. Together, these techniques allow the model to better capture the temporal continuity of animal behavior while keeping the simplicity of an ICL-based framework. We show an example of this process for labeling one second in Figure \ref{fig:ar_icl_pipeline}.

\begin{figure}[!ht]
    \centering
    \includegraphics[width=\linewidth]{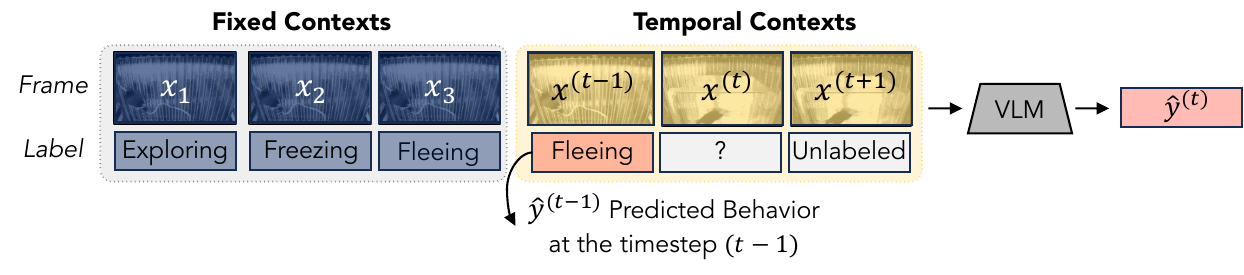}
    \caption{
    Proposed AR-ICL method for behavioral labeling at the $(t)$-th frame. 
    }
    \label{fig:ar_icl_pipeline}
\end{figure}

\subsubsection{Analysis-read Calcium Imaging Data}

In addition to behavioral labeling, we explored whether VLMs could assist with extracting binary matrices from calcium imaging video data, where zero represents when a neuron is inactive and one represents when a neuron is active. Unlike behavioral data, calcium imaging videos contain subtle and fine-grained florescent signals that reflect neural activity at cell level resolution. Current methods to extract this information involve a piecemeal process of various scripts and libraries that is cumbersome and timely to execute. Using a VLM could potentially offer a one-step solution to this process.

We frame this as a coarse neural classification task, where each calcium imaging clip is labeled based on the visually discernible fluorescent activity (i.e., if something lights up in a given region). We also adopt an ICL framework, providing the VLM with a small set of example calcium imaging clips paired with expert annotations. 

Due to limitations in how much detail VLMs can understand from a video, instead of asking the VLM to reason over the full-resolution clip, we decided to partition each frame into a small grid and represent activity in a binary manner. To do this, we use an $n\times n$ grid over the full frame. If any noticeable fluorescent activity occurs within a quadrant, that region is labeled as active ($1$), otherwise it is labeled as inactive ($0$). This gives us a compact binary matrix representation per second that captures coarse spatial structure while ignoring fine-grained cellular details that are difficult for current VLMs to interpret reliably. Unlike the behavioral task, we do not use AR-ICL here, instead treating each calcium second independently. This is because, unlike the behavioral task, calcium fluorescence activity is much noisier and does not transition as clearly across time. As a result, we don't believe that providing the previous prediction will improve performance, and in fact may be worse for this task. 

\subsection{AI-enhanced Tensor Analysis}
After data preparation, 
we obtain neuron data with three modes (trials, time, neuron) represented as $\T{X} \in \mathbf{R}^{I \times J \times K}$ and behavior data with three modes (trials, time, behavior type) represented as 
$\T{Y} \in \mathbf{R}^{I \times J \times M}$.
These data are naturally represented as tensors which exhibit multi-way patterns between trials, times, neurons, and behavior types. 
Tensor decomposition is a fundamental analysis tool to discover latent patterns of multi-way patterns in tensors, and has been actively used in neuroscience~\cite{acar2013understanding,acar2014structure,zhou2016linked,hunyadi2016power}.
Among various tensor decomposition models, CANDECOMP/PARAFAC decomposition (CPD)~\cite{kolda2009tensor} has been effectively used due to its simplicity and interpretability~\cite{kolda2009tensor}.
Given neuron data $\T{X} \in \mathbf{R}^{I\times J \times K}$ and rank $R$, 
CP decomposes $\T{X}$ into sum of $R$ rank-one components as follows.
\begin{equation}
\T{X} \approx \sum_{r=1}^{R}\lambda_r \vect{a}_r \circ \vect{b}_r \circ \vect{c}_r
            = \llbracket {\boldsymbol{\lambda}} ; \mat{A}, \mat{B}, \mat{C} \rrbracket \Leftrightarrow x_{ijk} = \sum_{r=1}^{R}\lambda_r a_{ir} b_{jr} c_{kr}, 
\end{equation}            
where the $r$th rank-one component $\vect{a}_r \circ \vect{b}_r \circ \vect{c}_r$ 
consists of distinct latent patterns between different modes, and $\circ$ denotes an outer product.
Here, $\mat{A} \in \mathbf{R}^{{I \times R}}, \mat{B} \in \mathbf{R}^{{J \times R}}$, and $\mat{C} \in \mathbf{R}^{{K \times R}}$ are factor matrices corresponding trial, time, and neuron, and $\boldsymbol{\lambda} \in \mathbf{R}^{R}$ denotes a weight vector of components. 
Element-wise, an each tensor entry $x_{ijk}$ can be reconstructed with
factor vectors of $\vect{a}_{i,:}, \vect{b}_{j,:}, \vect{c}_{k,:}$
$a_{ir}$ denotes the $r$-th factor values of the $i$-th index of the first mode.

When there are tensors where modes are coupled, coupled tensor decomposition has been used to integrate information comprehensively~\cite{acar2013understanding,acar2014structure}.
%
Given neural data $\T{X} \in \mathbf{R}^{I\times J \times K}$, behavior data $\T{Y} \in \mathbf{R}^{I\times J \times K}$, and rank $R$, where trial and time modes are coupled, 
CP decomposes $\T{X}$ and $\T{Y}$ into factor matrices $\{\mat{A}, \mat{B}, \mat{C}, \mat{D}\}$ as follows:
\begin{equation}
     L = ||\T{X} - \sum_{r=1}^{R} \lambda_r \bar{\vect{a}}_r \circ \bar{\vect{b}}_r \circ \bar{\vect{c}}_r ||_{F}^2 +
         ||\T{Y} - \sum_{r=1}^{R} \gamma_r \bar{\vect{a}}_r \circ \bar{\vect{b}}_r \circ \bar{\vect{d}}_r||_{F}^2,
\end{equation}
%
where $\mat{A}$ and $\mat{B}$ are shared factor matrices for the trial and time modes, respectively, 
and $\mat{C}$ and $\mat{D}$ are neuron and behavior factor matrices.
We impose a non-negativity constraint on the factor matrices to facilitate interpretability, such that $\bar{\vect{a}} = \phi(\vect{a})$ where $\phi$ is a non-negative activation function.
Moreover, $\boldsymbol{\lambda}$ and $\boldsymbol{\gamma}$ are trained to weight
components, with larger values indicating stronger contributions, which enables direct identification of shared common and tensor-specific components.

Recently, Neural Additive Tensor Decomposition (NeAT) has been proposed to enhance decomposing tensors~\cite{ahn2024neural}, by applying an individual neural network to each component. Thus, it maintains the interpretability of CP decomposition while capturing non-linear patterns.
%
Given neural data $\T{X}$, behavior data $\T{Y}$, and rank $R$, 
NeAT decomposes $\T{X}$ and $\T{Y}$ into factor matrices $\{\mat{A}, \mat{B}, \mat{C}, \mat{D}\}$ as follows:
{\small
\begin{equation}
  \begin{split}
     L =
     \sum_{\alpha=(i, j, k)}\Big(x_{ijk} - \sum_{r=1}^{R} f_r(\phi([{a}_{ir}, {b}_{jr}, c_{kr}]))\Big)^2 + \\
     \sum_{\beta=(i, j, l)}\Big(y_{ijl} - \sum_{r=1}^{R} g_r(\phi([{a}_{ir}, {b}_{jr}, d_{lr}]))\Big)^2,
    \end{split}
\end{equation}
}
where $f_r, g_r$ are multi-layer perceptrons (mlp) for each tensor, and 
[,] denotes a concatenation, e.g.,$[{a}_{ir}, {b}_{jr}, c_{kr}] \in \mathbf{R}^{3}$. 
Similar to CPD, NeAT maintains the additivity of components similar to CPD, which makes it easy to interpret each component separately to other components, unlike other types of neural tensor decomposition~\cite{liu2019costco}.
For coupled tensors, we share factor matrices $\mat{A}$ and $\mat{B}$ but have separate mlps to learn tensor-specific interactions.
To identify shared and unshared components,
we use one-layer mlp and calculate the sum of weights and bias in mlp. 
Since we also impose non-negativity on factor matrices, naturally the larger values of weights in mlps indicate stronger contributions of the corresponding components. 

\subsection{AI-driven Pattern Interpretation}

A central goal of our pipeline is to move beyond automated pattern extraction, and close the loop by moving toward assisted scientific interpretation. After latent tensor decomposition reveals latent patterns between neural and behavior components, domain experts need to then determine what these patterns mean and represent. Like the other traditional methods we've mentioned, this step is often time-consuming and requires familiarity with what latent factors represent as well as how they relate to different neuroscience interpretations. We therefore design a system that can automatically generate plausible, literature grounded hypothesis for each component discovered, aiding experts by providing them a nice direction for where to start, serving as an interpretive aid.

Recent advances have shown that both LLMs and VLMs have demonstrated strong capabilities in both scientific reasoning and hypothesis generation across multiple domains \cite{song2025evaluatinglargelanguagemodels}. These models can combine visual information from factor plots with relevant domain knowledge to produce explanations in natural language rather than raw numerical output. This is useful because latent tensor factors are often abstract and not always straightforward to interpret, allowing VLMs to serve as a practical bridge between outputs and interpretation.

For each latent component, the model is provided with (1) the \textbf{visualization} of the component’s trial-wise factor plots, (2) a small set of in-context examples consisting of previously interpreted components with \textbf{expert annotations}, and (3) \textbf{relevant neuroscience literature}.

To familiarize the VLM with this task, we again utilize ICL to provide a small set of example latent factor plots paired with an expert written interpretation and a discovery score between $1-5$ \cite{NEURIPS2020_1457c0d6}. The discovery score reflects how strongly the observed pattern is supported by existing neuroscience literature, where lower scores indicate weak or no prior support and higher scores indicate strong prior support. Including the discovery score encourages the model to reason not only about what a pattern may represent, but also how confident the field should be in that interpretation based on prior evidence. These examples show the VLM the expected level of detail, tone, and evaluation criteria which allows the model to generate properly formatted hypothesis for new components without additional fine-tuning.

To further anchor the models outputs in established neuroscience findings, we also leverage retrieval-augmented generation (RAG) \cite{gao2024retrievalaugmentedgenerationlargelanguage, lewis2021retrievalaugmentedgenerationknowledgeintensivenlp}. In addition to the ICL examples, the model also receives related neuroscience papers that analyze related latent patterns \cite{gonzalez2020role, giustino2015role, barrett2018prefrontal, sangha2013safety, sierra2011dissociable, yan2018coding, jeffery2004proposed}. This provides additional background context about how similar patterns have been interpreted within the literature, encouraging interpretations that are grounded, while still allowing room for novel but plausible hypothesis and ideas.

This system is essentially able to act as a hypothesis generation assistant, whose role is to propose plausible interpretations that can guide analysis, particularly in cases where latent factors are difficult to interpret directly. Because latent factorization methods are used across many scientific domains, this framework may be generalizable to other settings where latent factor analysis can reveal patterns which are not initially obvious from raw data \cite{Sidiropoulos_2017}.

\section{Evaluating the Pipeline}
We evaluate data preparation and tensor analysis compared to existing baselines.
\subsection{ICL Labeling Evaluation}
\subsubsection{Behavioral Video Labeling Evaluation} \label{sec:exp:behavior}
To evaluate our approach to behavior annotation, as well as the gains introduced by AR-ICL, we conduct a series of experiments on the second-by-second behavior classification task. For these experiments, we use the recently released Qwen3-VL-32B-Instruct VLM \cite{qwen3technicalreport, Qwen-VL, Qwen2VL, Qwen2.5-VL}. 
This is because it is open-source supporting
reproducibility and ease of access, is relatively low cost for domain experts to use, and demonstrates state-of-the-art performance among open-source VLMs.

As a transformer-based baseline, we use DINOv2 to embed the same ICL example frames used for the VLM and perform nearest-neighbor classification for each new one-second segment based on embedding similarity to the labeled examples. This provides a strong visual representation learning baseline and is similar to past approaches \cite{goulart2025preliminaryusevisionlanguage, caron2021emergingpropertiesselfsupervisedvision}.

In order to see whether the improvements from AR-ICL are due specifically to the autoregressive label of previous predictions, rather than simply providing additional temporal context, we also evaluate a temporal ICL variant where the model is given the previous and next one-second segments without any label. This allows us to isolate the effect of the autoregressive component and understand how much temporal context alone contributes beyond the original ICL examples. Additionally, we also include results from just ICL examples. Across all configurations, we use the same $3$ randomly selected ICL examples, one for each behavior class.

We evaluate all methods on a total of 3,240 one-second consecutive video segments. The behaviors are distributed as follows: Freezing: $410 / 3,240$ ($12.7\%$), Fleeing: $21 / 3,240$ ($0.6\%$), and Grooming/Exploring: $2,809 / 3,240$ ($86.7\%$). As shown, the dominant class consists of safe (non-fear) behaviors such as grooming and exploring, which we group together into a single class due to their high visual similarity and because distinguishing fear-related behaviors is of greater scientific importance. Importantly, the fear-related behaviors (freezing and especially fleeing) are much rarer, making them more difficult to learn and reliably detect, even for human annotators, despite being critical behaviors in the dataset.

Due to this heavy class imbalance, we report multiple evaluation metrics. We report macro F1 across all three classes, balanced accuracy, and Matthews correlation coefficient (MCC), which has been shown to provide more reliable and informative evaluation than standard F1 in highly imbalanced datasets \cite{chicco2020mcc}. These metrics are shown in Table \ref{tab:overall_metrics_behavior}. Detailed per-class precision, recall, F1, and F2 scores are provided in the Appendix.

As shown in Table \ref{tab:overall_metrics_behavior}, ICL on its own has a dramatic impact on VLM performance for this task. Without ICL, the model fails to identify any instances of fleeing and performs poorly overall, largely due to over-predicting the dominant class. In fact, the no ICL VLM performs not much better than the DINOv2 nearest-neighbor baseline, which has no reasoning capabilities. We observed this qualitatively as well, when the VLM was asked to describe the clips, it often failed to even recognize there was 
a mouse present.

Introducing standard ICL leads to a substantial improvement in overall performance, indicating that in-context examples provide crucial task grounding for the model. Providing temporal context alone, without autoregressive feedback, does not yield additional gains and in fact slightly degrades performance relative to standard ICL. In contrast, our proposed AR-ICL method consistently achieves the best performance. AR-ICL achieves the best 
across all the
metrics, including macro F1, balanced accuracy, and MCC.

\subsubsection{Calcium Imaging Video Labeling Evaluation}
We also conducted a preliminary evaluation of ICL for labeling calcium imagine videos. As mentioned before, calcium imaging data contains subtle fluorescence signals at very small scales, making this a signifigantly more challenging than behavioral labeling.
To study this, we partition each frame into an $n \times n$ grid and ask the VLM to predict a binary matrix with each entry corresponding to the respective region. We evaluate several resolutions including: $2 \times 2$, $3 \times 3$, and $5 \times 5$. From this we observe that as the grid becomes finer, the performance drops as shown in Figure \ref{fig:calcium_resolution_performance}. This indicates the the VLM can capture low resolution patterns which are indicated by the relatively high F1 score and accuracy. However we can see that as the resolution increases, the performance drops.
Therefore we view this component as an exploratory step rather than a replacement for current methods. 

\begin{figure}[!ht]
    \centering
    \includegraphics[width=0.35\textwidth]{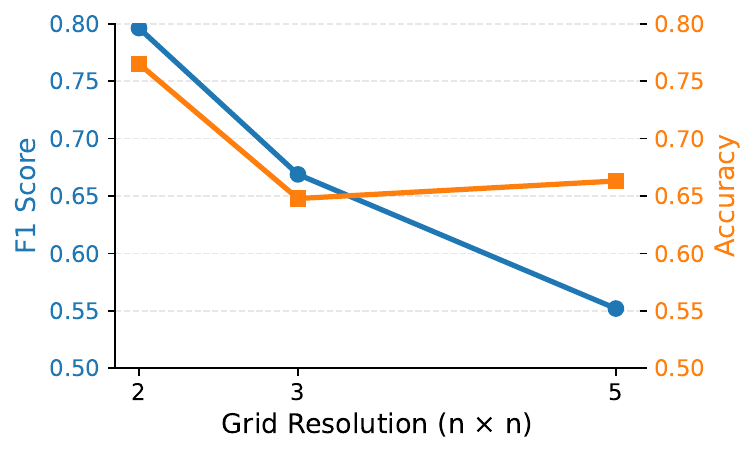}
    \caption{
    Performance of VLM-based calcium activity. F1 score (left axis) and accuracy (right axis).
    }
    \label{fig:calcium_resolution_performance}
\end{figure}

\begin{table}[h]
\centering
\small
\caption{Overall performance comparison. Best values per column are bolded.}
\label{tab:overall_metrics_behavior}
\begin{tabular}{lccc}
\toprule
\textbf{Method} & \textbf{Macro F1} & \textbf{Balanced Acc} & \textbf{MCC} \\
\midrule

DINO Baseline & 0.203 & 0.441 & 0.128 \\
Qwen3 – No ICL & 0.370 & 0.365 & 0.130 \\
Qwen3 – ICL & 0.492 & 0.782 & 0.451 \\
Qwen3 – Temporal ICL & 0.490 & 0.777 & 0.424 \\
Qwen3 – AR-ICL & \textbf{0.545} & \textbf{0.801} & \textbf{0.517} \\

\bottomrule
\end{tabular}
\end{table}

\subsection{Tensor Decomposition} \label{sec:exp:tensor}
We select two tensor decomposition model: CPD and NeAT for interpretability. 
We evaluate tensor decomposition results using Root Mean Squared Error (RMSE)
defined as follows: $\sqrt{\frac{1}{N}\sum_{\alpha}(x_{\alpha} - \tilde{x}_{\alpha})^2}$, where $\tilde{x}$ is a reconstruction entry corresponding to index $\alpha$ and $N$ is the total number of tensor entries. 
We have seven subjects/mice in total where neuron tensor includes three modes of (trial $\times$ time $\times$ neuron) where its
size is ($33 \times 6000 \times N_s$) and $N_s$ ranging from 948 to 2339.
Behavior tensor includes two modes of (trial $\times$ time) where its size is
($33 \times 6000$). 
In our dataset, trial and time modes are coupled.
The value of behavior matrices are binary where
0 indicates safety (exploring state) and 1 indicates fear (freezing and fleeing). 
Since the most of activities are exploring, we sample zero values while we use all freezing and fleeing activities. 
We split tensors and coupled matrices into training and test datasets with 9:1 ratios.
\Cref{tab:tensor_methods} shows the test RMSE of CPD and NeAT decomposition on non-coupled (used only neuron tensor) and coupled tensors (used both neuron and behavior tensors) on two subjects. Note that results are provided in the Appendix. 
NeAT shows the lowest test RMSE compared to CPD since it can capture nonlinear patterns, and 
effectively decomposes couple tensors by using separate mlp for different tensors.

\begin{table}[h!]
\centering
\caption{Test RMSE of tensor completion. Note that C and NC indicate if tensors are coupled or not.
}
\label{tab:tensor_methods}
{\small
\begin{tabular}{l l cccc}
\toprule
& & \multicolumn{2}{c}{CPD}  & \multicolumn{2}{c}{NeAT}  \\
\cmidrule(lr){3-4} \cmidrule(lr){5-6} 
\multicolumn{2}{c}{\textbf{Subjects}}
& \multicolumn{1}{c}{NC} & \multicolumn{1}{c}{C}  
& \multicolumn{1}{c}{NC} & \multicolumn{1}{c}{C}  \\
\midrule
     & 1 & 0.1457 & 0.2077 & \textbf{0.1099} & 0.1113 \\
     & 2 & 0.1255 & 0.1483 & 0.0937 & \textbf{0.0919} \\
\bottomrule
\end{tabular}
}
\end{table}
\section{Evaluating Interpretation \& Discovery}
\label{sec:discovery}
Before evaluating the full end-to-end pipeline, we first assess the AI-driven interpretation in isolation. Specifically, we evaluate how well the VLM can interpret expert-analyzed latent components by comparing model-generated hypotheses to expert-written descriptions and discovery scores. This allows us to measure agreement in both the scientific reasoning and the strength of literature support attributed to each latent pattern.
We then demonstrate the utility and performance of our proposed end-to-end pipeline on a small dataset of mouse behavior and calcium imaging data for two subjects during a single trial.

\subsection{Research Question}
Mice were trained to discriminate between different stimuli presented during a single trial and we hypothesize that there will be unique populations of neurons that are highly co-engaged when the animal engages in specific behaviors and during distinct presentations of the different stimuli. Additionally, if no stimuli are presented during the trial, we anticipate no discernible pattern of neuronal activity to be detected.

\subsection{Experimental Setup \& Data Collection}
Mice underwent a fear conditioning and discrimination behavioral training paradigm. Three different environments, as seen in Figure 6, were used and include: a safe environment (CS-) that should signal safety, a threatening environment (CS+) that signals threat, and a home cage analog environment (NS) that is familiar to the mouse. Mice were first habituated to the different stimuli and environments to ensure no inherent fear or preference for any environments or cues.  Next, they were fear conditioned in the CS+ environment, which involves three sequential presentations of mild foot shocks paired with an auditory stimulus (upsweep tone). The mice were then exposed to each of the three different environments everyday for eight days.
Behavioral data and calcium imaging data was collected during each of the trials on each of the days, except for fear conditioning day where no calcium imaging data was collected.

\begin{table*}[!ht]
\centering
\caption{Examples of AI-driven interpretation of latent neural components. We compare expert-written interpretations with VLM-generated hypotheses and discovery scores.}
\label{tab:discovery_examples}
\small
\begin{tabular}{p{0.6cm}>{\raggedright\arraybackslash}p{6.8cm} >{\raggedright\arraybackslash}p{7.2cm} c c}
\toprule
\textbf{Comp.} & \textbf{Expert Interpretation} & \textbf{VLM Interpretation} & \textbf{Expert} & \textbf{VLM} \\
 &  &  & \textbf{Score} & \textbf{Score} \\
\midrule
1 &
\colorbox{Lavender}{High neuron coactivity during the threatening and familiar} \colorbox{Lavender}{environments}. Unexpected result, but might be due to (1) the mouse knows the familiar environment poses no threat, and (2) the threatening environment has a strong negative valence associated with it — where the animal is afraid (demonstrating high freezing behavior) throughout learning to discriminate the environments. &
\colorbox{Lavender}{High neuron co-activity during} early learning phase and \colorbox{Lavender}{sustained activity throughout the learning process}, with peaks occurring in both safe and threatening environments, suggesting a general engagement during environmental exploration and initial learning. This latent variable is not directly supported by previous research, but some studies report broad prelimbic cortex activation during early learning stages.&
1 & 2 \\

\addlinespace
\midrule
\addlinespace

2 &
\colorbox{Lavender}{Threatening environment responding component, with} 
\colorbox{Lavender}{high neuron co-activity during threatening environment} 
\colorbox{Lavender}{exposure}; especially near the end of the behavior paradigm, where the mouse knows the threatening environment signals threat. &
\colorbox{Lavender}{High neuron co-activity during the threatening environment}, with a peak occurring near the end of learning, suggesting increased engagement as the animal consolidates threat-related memory. &
4 & 4 \\

\addlinespace
\midrule
\addlinespace

3 &
\colorbox{Lavender}{High neuron co-activity during familiar environment with} 
\colorbox{Lavender}{some activity in the safe and threatening environments} towards the end of learning (suggesting mouse is recognizing the environments). &
\colorbox{Lavender}{High neuron co-activity during the initial phase of learning} \colorbox{Lavender}{and again during the later phase of learning}, with minimal activity during the middle phase. This suggests a biphasic engagement pattern, possibly reflecting early encoding and late consolidation or reactivation processes. &
3 & 3 \\
\bottomrule
\end{tabular}
\end{table*}

\subsection{Evaluating AI-driven Interpretation}
\subsubsection{Agreement with Expert Interpretations}
To evaluate the effectiveness of our AI-driven interpretation module, we assess how well the VLM can generate plausible hypotheses for latent components. Instead of just measuring prediction accuracy, this evaluation focuses more on the interpretive alignment between the model generated explanations and expert explanations, as well as the agreement scores that quantify the amount of existing literature support for each pattern. 

For each latent component, the VLM was provided five ICL examples of previously interpreted explanations and discovery scores. In addition, the model received contextual background through RAG using a related neuroscience study that describes similar latent neural patterns. This setup is intended to mirror real world use of the system involving a domain expert supplying a small number of labeled analyzed examples as well as relevant literature, then having the model assist in interpreting newly discovered components.

In total we evaluated twelve components. For each component, we compared the VLM generated interpretation to the expert's written explanation and compared the predicted discovery socre (1-5) to the expert discovery score. Agreement on discovery scores was measured using quadratic-weighted Cohen's kappa in order to account to the scale, penalizing larger disagreements more heavily than smaller ones (i.e.,  1 vs 5 is worse than 3 vs 5).

Across all components, the model achieved a weighted \textbf{Cohen's kappa of 0.59}, which indicates moderate agreement between the VLM and domain experts' \cite{McHugh2012Kappa}. Importantly, most disagreements were off by only one (e.g. predicting a $2$ when the expert assigned a $1$), rather than large mismatches. This suggests that the model was generally able to capture how well supported the implications of latent factors were from prior literature. 

The VLM's interpretations frequently aligned with experts at the level of identifying which behaviors, or learning phases corresponded with threat related or safe related activity. In cases where the model's description differed from the domain experts', it often proposed a broader hypothesis, while the experts provided a more detailed hypothesis, linking interpretations to the environment. Examples of these comparisons are shown in Table \ref{tab:discovery_examples}, where we provide some examples of expert and VLM interpretations along with their corresponding discovery scores.

Overall however, these results support the role of the VLM as an automated latent factor interpreter and hypothesis generation assistant. The model consistently demonstrates the ability to translate abstract latent factors into coherent interpretations that generally align with expert reasoning. While not as detailed, these outputs provide a starting point that can help guide experts, speeding up the analysis of newly discovered patterns.
\subsubsection{Full-Pipeline Discovery Case Study}
To demonstrate our full pipeline, we applied it end-to-end on two subjects during a single trial. Behavioral labels were obtained automatically using our VLM annotation piece of the pipeline, and these labels were combined with calcium imaging data in our tensor analysis stage to extract latent components. These components were then analyzed by both a domain expert and VLM.
It's important to note that these components were not previously analyzed by experts, which allows us to assess how well our pipeline can assist in generating plausible scientific hypothesis that accelerate discovery.

For Subject 1, both the expert and the VLM characterized the component as largely non-specific and highly variable across the trial. The expert noted a possible late increase in activity but assigned a Discovery Score of $1$. The VLM described the pattern as noisy and lacking clear behavioral structure, and also assigned a Discovery Score of $1$. This agreement indicates that the VLM appropriately recognized the absence of strong literature-supported structure in the factor.
For Subject 2, the expert observed periods of elevated co-activity that might relate to behavior such as freezing or grooming, assigning a Discovery Score of 3. The VLM identified multiple sharp peaks distributed across the session without clear alignment to specific task epochs and assigned a Discovery Score of 2. Although slightly lower, the VLM’s interpretation was directionally consistent with the expert’s assessment.
Overall, the VLM-generated hypotheses was pretty close when compared to expert interpretations in both description and discovery score. These results suggest that, when used within the full pipeline, the VLM can assist in forming plausible interpretations of newly extracted components. This can be useful in helping domain experts more quickly analyze latent factors and come up with valid hypothesis.

\vspace{1mm}
\vspace{-4mm}
\section{Related Work}
Similar works have recently proposed models that aim to optimize experimental research workflows through the use of AI-enhanced methods. Gottweis and colleagues proposed a revolutionary multi-agent system intended to serve as an "AI co-scientist" that can assist researchers in developing hypotheses, proposals, and experimental designs \cite{gottweis2025towards}. For example, LLMs can be used to assist in the development of experimental hypotheses and interpretation of results \cite{song2025evaluating}, and further visual inputs into these models can minimize limitations on the data that scientists can generate and analyze.

One important aspect of our work is to strive to include and empower the domain expert, rather than replacing them altogether, and as such, we identified In-Context Learning as an important paradigm that is useful in that goal. The work of Agile Modeling by Stretcu et al. \cite{stretcu2023agile} shares a similar goal of building a system which empowers a domain expert to leverage machine learning models in their work, and have demonstrated the utility of this framework in the context of conservation and ecology \cite{dumoulin2025search}.

\section{Conclusions}
We propose an AI-enhanced pipeline for behavior neuroscience, which automates existing time-consuming and tedious workflows while providing an easy-to-use interface to cutting-edge foundation models for domain experts. 
Our proposed AR-ICL framework for automated behavioral labeling achieves a Macro F1 score of $0.545$, a balanced accuracy of $0.801$, and an MCC of $0.517$ on extremely imbalanced and hard to label behavioral data. 
We enhance tensor analysis by leveraging an interpretable neural tensor decomposition model, showing $46\%$ lower test RMSE than CPD. 
Finally, we interpret tensor analysis results using a RAG \& ICL enhanced VLM, whose outputs are consistent with domain experts explanation and achieve similar discovery scores. We believe that our proposed pipeline is a significant step toward transforming behavioral neuroscience research.

\section{Limitations and Ethical Considerations}
\noindent{\bf Limitations}
On the application side, 
lens implantation into the region of interest can cause local tissue damage, and chronic imaging across extended time periods can cause alterations to lens alignment and diminished quality in optical recording of neuronal activity. Also, although the use of calcium binding as an indicator of neuronal activity is supported and accepted in the field \cite{wei2020comparison, resendez2016visualization}, genetically encoded calcium indicators have slow decay kinetics, which constrains the temporal precision of the neural dynamics that are aligned with mouse behavior outputs.

On the algorithmic side, the most significant limitation is that ICL, as currently used and adapted, does not perform as well for calcium imaging data preparation compared to behavior extraction. We are actively investigating how to improve performance while maintaining the spirit of the approach, which requires zero to minimal model manipulation knowledge from its end-user.

\noindent{\bf Ethical Considerations} The University of California, Riverside Institutional Animal Care and Use Committee approved all procedures following the NIH guidelines for the care and use of laboratory animals. 
%




\newpage
\section{GenAI Disclosure}
GenAI has been used only for purposes of improving writing and figure plotting and generation. All content and ideas are original and any such use of  GenAI has been carefully vetted and approved.
Finally, our proposed method integrates GenAI models such that some of the results are generated by those models, however, this is part the entire framing of the work. 

\balance
\bibliographystyle{ACM-Reference-Format}
\bibliography{BIB/base,BIB/dawon,BIB/jordan,BIB/vagelis}


\appendix
\onecolumn

\section{Per-Class Behavioral Labeling Performance}

Here, we provide a more detailed breakdown of model behavior under heavy class imbalance, we report per-class precision, recall, F1 score, and F2 score in Table \ref{tab:behavior_per_class}.

As shown in Table \ref{tab:behavior_per_class}, ICL on its own has a dramatic impact on VLM performance for this task. Without ICL, the model fails to identify any instances of fleeing and achieves an F1 score of only 0.161 for freezing. Although the no ICL model achieves a high F1 score for grooming/exploring of 0.914, we believe that this is attributed to over-predicting the dominant class. Interestingly, the DINOv2 baseline achieves higher recall for freezing and fleeing than the VLM without ICL, suggesting that the VLM struggles to interpret these out-of-distribution experimental videos without specific context.

Introducing standard ICL leads to a substantial improvement across all fear-related behaviors. While the F1 score for grooming/exploring decreases slightly, this is expected under class imbalance when the model becomes more sensitive to minority classes. Both freezing and fleeing detection improve considerably, indicating that in-context examples provide crucial task grounding for the model. It is also important to note that fleeing is an extremely rare behavior (only $0.6\%$ of the data), meaning that even a small number of false positives can significantly reduce the F1 score. Despite this, we observe relatively high recall for fleeing when using ICL-based methods, suggesting that when fleeing does occur, the model is often able to correctly detect it.

Providing temporal context alone, without autoregressive feedback, does not yield additional gains relative to standard ICL. In contrast, our proposed AR-ICL method achieves the strongest performance on the minority fear-related behaviors, yielding the highest F1 scores for both freezing and fleeing while maintaining strong performance on grooming/exploring.

\begin{table*}[h]
\centering
\caption{Per-class performance on behavioral video labeling. Best values per column are bolded.}
\label{tab:behavior_per_class}
\begin{tabular}{lcccccccccccc}
\toprule
& \multicolumn{4}{c}{Freezing} 
& \multicolumn{4}{c}{Fleeing} 
& \multicolumn{4}{c}{Grooming/Exploring} \\
\cmidrule(lr){2-5} \cmidrule(lr){6-9} \cmidrule(lr){10-13}
\textbf{Method} 
& Precision & Recall & F1 & F2
& Precision & Recall & F1 & F2
& Precision & Recall & F1 & F2 \\
\midrule

DINOv2 Baseline
& 0.251 & 0.612 & 0.356 & 0.475
& 0.007 & 0.571 & 0.013 & 0.031
& 0.930 & 0.138 & 0.241 & 0.166 \\

Qwen3 – No ICL
& 0.306 & 0.144 & 0.196 & 0.161
& 0.000 & 0.000 & 0.000 & 0.000
& 0.878 & \textbf{0.952} & \textbf{0.914} &\textbf{0.936} \\

Qwen3 – ICL
& 0.422 & \textbf{0.941} & 0.583 & 0.755
& 0.042 & 0.714 & 0.080 & 0.171
& \textbf{0.986} & 0.692 & 0.813 & 0.736 \\

Qwen3 – Temporal ICL
& 0.484 & 0.917 & 0.634 & 0.778
& 0.027 & \textbf{0.762} & 0.052 & 0.119
& 0.980 & 0.653 & 0.784 & 0.700 \\

Qwen3 – AR-ICL
& \textbf{0.529} & 0.905 & \textbf{0.668} & \textbf{0.792}
& \textbf{0.051} & 0.714 & \textbf{0.096} & \textbf{0.199}
& 0.980 & 0.784 & 0.871 & 0.817 \\

\bottomrule
\end{tabular}
\end{table*}

\section{Prompts Used}

Here we show the prompts used for both the behavior task as well as the AI-driven interpretation task. The AR-ICL prompt for the behavior task is shown in Figure \ref{fig:aricl_prompt}, the AI-driven interpretation prompt is shown in Figure \ref{fig:discovery_prompt}, and the ICL prompt for the calcium imaging video task is shown in Figure \ref{fig:calcium_icl_prompt}.

\begin{figure}[!ht]
\centering
\fcolorbox{blue!50!black}{blue!5!white}{
\begin{minipage}{0.97\linewidth}

\textbf{Task:}  

Label the mouse’s behavior in this n-second clip.  
Return exactly n label(s) in a Python list \texttt{[l1, ..., ln]}.

Allowed: \texttt{Freezing}, \texttt{Fleeing}, \texttt{Exploring}.  
\texttt{Freezing} = absolutely no visible movement across the whole second (no head/ear/whisker/tail or body motion).  
\texttt{Exploring} = any visible movement that isn’t fast fleeing; includes slow stepping in place, head/whisker/ear/tail motion, sniffing, re-orienting, or brief rearing with little displacement.  
\texttt{Fleeing} = fast, sustained locomotion, large displacement, motion blur, or dashing.  

If unsure between fleeing and exploring, choose fleeing if movement is more rapid.  
Rules: if any movement is seen in any frames, do NOT output \texttt{Freezing}.  
Output only the list.

\vspace{0.35cm}
\textbf{Examples (Fixed ICL):}

\begin{tabular}{l c l}
Video/Frames 1: &
\includegraphics[width=2.2cm,height=1.3cm]{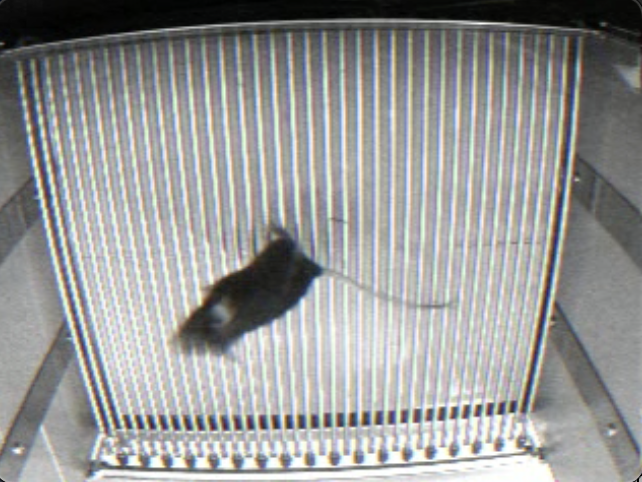} &
$\rightarrow$ \texttt{[Exploring]} \\

Video/Frames 2: &
\includegraphics[width=2.2cm,height=1.3cm]{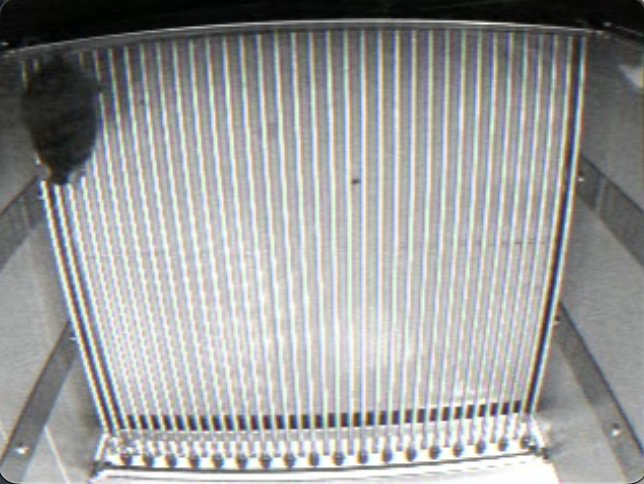} &
$\rightarrow$ \texttt{[Freezing]} \\

Video/Frames 3: &
\includegraphics[width=2.2cm,height=1.3cm]{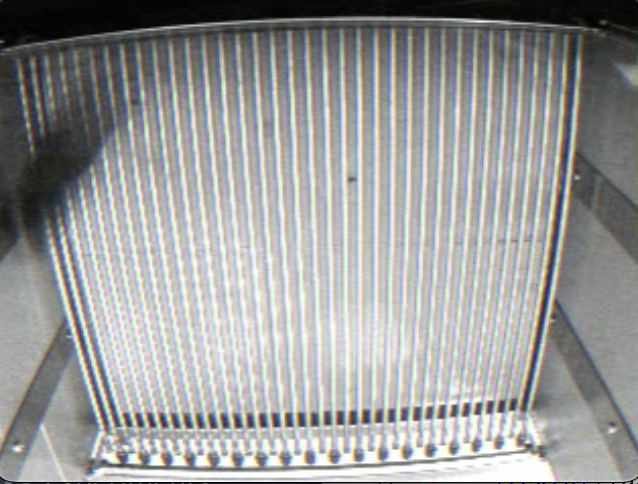} &
$\rightarrow$ \texttt{[Fleeing]} \\
\end{tabular}

\vspace{0.55cm}
\textbf{Temporal Context (AR-ICL):}

\textit{The following frames correspond to the immediately preceding second.  
Use this temporal context to inform your decision, while basing your label on the current frames.}

\begin{tabular}{l c l}
Previous Second ($x_{t-1}$): &
\fcolorbox{black!60}{gray!15}{
\parbox[c][1.3cm][c]{2.2cm}{\centering $\mathbf{x_{t-1}}$}
} &
$\rightarrow$ \texttt{[$\hat{y}_{t-1}$]} \\
\end{tabular}

\vspace{0.4cm}

\textit{The following frames correspond to the immediately next second.  
Use this temporal context to inform your decision, while basing your label on the current frames.}

\begin{tabular}{l c}
Next Second ($x_{t+1}$): &
\fcolorbox{black!60}{gray!15}{
\parbox[c][1.3cm][c]{2.2cm}{\centering $\mathbf{x_{t+1}}$}
} \\
\end{tabular}

\vspace{0.55cm}
\textbf{Target (Current Second $x_t$):}

\begin{tabular}{l c l}
Target Frames: &
\fbox{
\parbox[c][1.3cm][c]{2.2cm}{\centering \textbf{Input}}
} &
$\rightarrow$ \texttt{[ \ \ ]} \\
\end{tabular}

\end{minipage}
}
\caption{AR-ICL prompt used for temporally consistent behavioral labeling. In addition to fixed ICL examples, the model receives the previous second with its predicted label and the next second as unlabeled temporal context.}
\label{fig:aricl_prompt}
\end{figure}

\begin{figure}[t]
\centering
\fcolorbox{blue!50!black}{blue!5!white}{
\begin{minipage}{0.97\linewidth}

\textbf{Task:}  

First, describe the observable pattern of neural engagement in this factor based only on the figure.  
Second, compare this pattern to findings in the provided neuroscience literature.  
Finally, provide a scientific interpretation and assign a \textbf{Discovery Score from 1 to 5} based on how strongly the literature supports this interpretation.

A Discovery Score of 5 should only be given if the literature provides direct evidence for this specific type of neural population response in similar behavioral paradigms. If support is indirect, debated, or comes from different brain regions or tasks, assign a lower score.

Do not copy the examples. Base your answer only on the visual pattern and the referenced literature. Format exactly like the examples.

\vspace{0.5cm}
\textbf{Retrieved Scientific Context (RAG Component):}

\textit{The following pages are retrieved from a neuroscience paper describing neural population dynamics in the prelimbic cortex during learning. Use this as scientific background when evaluating the component.}

\begin{tabular}{c c c c}
\fbox{\parbox[c][1.3cm][c]{2.1cm}{\centering Paper Page 1}} &
\fbox{\parbox[c][1.3cm][c]{2.1cm}{\centering Paper Page 2}} &
\fbox{\parbox[c][1.3cm][c]{2.1cm}{\centering Paper Page 3}} &
\fbox{\parbox[c][1.3cm][c]{2.1cm}{\centering Paper Page 4}}
\end{tabular}

\vspace{0.6cm}
\textbf{Examples (ICL for Scientific Interpretation):}

\begin{tabular}{l c l}
Example 1: & \includegraphics[width=2.2cm,height=1.3cm]{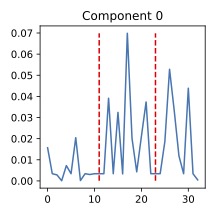} & $\rightarrow$ Expert Interpretation + Score 5 \\
Example 2: & \includegraphics[width=2.2cm,height=1.3cm]{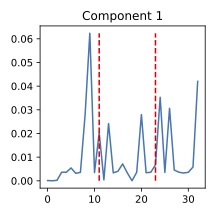} & $\rightarrow$ Expert Interpretation + Score 5 \\
Example 3: & \includegraphics[width=2.2cm,height=1.3cm]{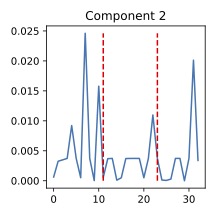} & $\rightarrow$ Expert Interpretation + Score 3 \\
Example 4: & \includegraphics[width=2.2cm,height=1.3cm]{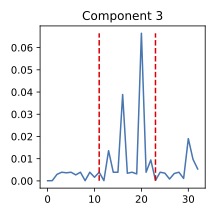} & $\rightarrow$ Expert Interpretation + Score 1 \\
Example 5: & \includegraphics[width=2.2cm,height=1.3cm]{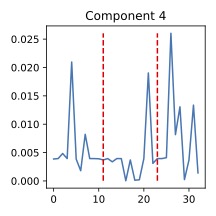} & $\rightarrow$ Expert Interpretation + Score 1 \\
\end{tabular}

\vspace{0.6cm}
\textbf{Target Latent Component:}

\begin{tabular}{l c l}
Component to Interpret: &
\includegraphics[width=2.2cm,height=1.3cm]{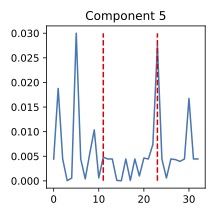} &
$\rightarrow$ \texttt{Model generates interpretation + Discovery Score}
\end{tabular}

\end{minipage}
}
\caption{Discovery ICL prompt used for AI-driven interpretation of latent neural components. The model receives retrieved neuroscience literature (RAG component), followed by several example latent factors paired with expert interpretations and discovery scores (ICL), and then a new latent component for analysis.}
\label{fig:discovery_prompt}
\end{figure}

\begin{figure}[!ht]
\centering
\fcolorbox{blue!50!black}{blue!5!white}{
\begin{minipage}{0.97\linewidth}

\textbf{Task:}  

Given this 1-second calcium imaging video segment, divide the frame into a $3 \times 3$ grid of equal-sized regions.  
Each cell corresponds to a fixed spatial region in the frame.  

For each region, determine whether there is visible calcium activity during this second.  
Neural activity is indicated by visible calcium fluorescence changes (e.g., brightening, transients, or sustained activation) relative to the local background.

Output a binary matrix where \texttt{1} indicates activity and \texttt{0} indicates no activity.  
Return exactly one matrix as a Python list of lists, e.g. \texttt{[[0,1,0],[...],...]]}.  
Output only the matrix.

\vspace{0.35cm}
\textbf{Examples (Fixed ICL):}

\begin{tabular}{l c l}
Example Clip 1: &
\includegraphics[width=2.2cm,height=1.3cm]{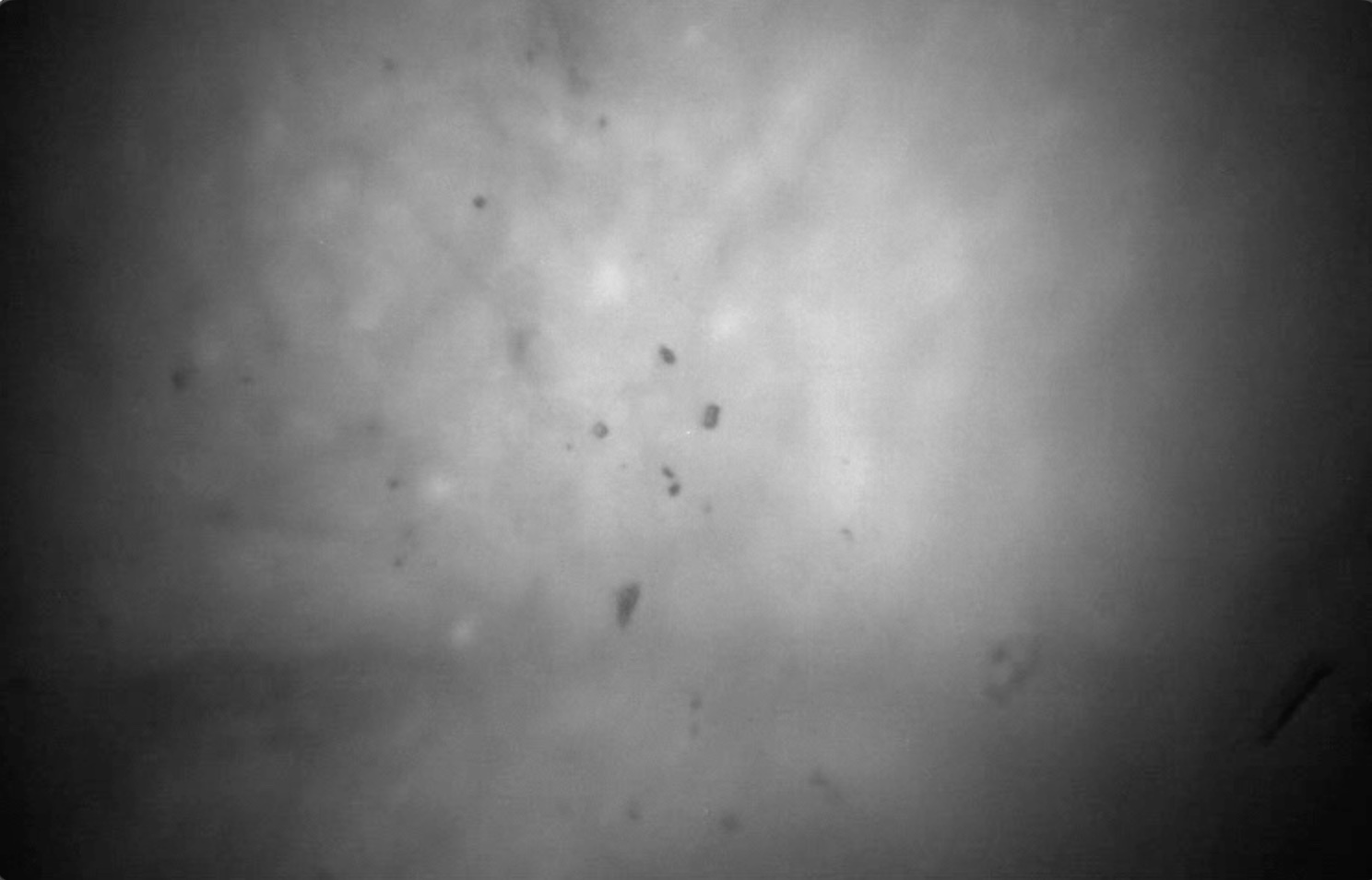} &
$\rightarrow$ \texttt{[[1,1,0],[1,1,0],[0,1,0]]} \\

Example Clip 2: &
\includegraphics[width=2.2cm,height=1.3cm]{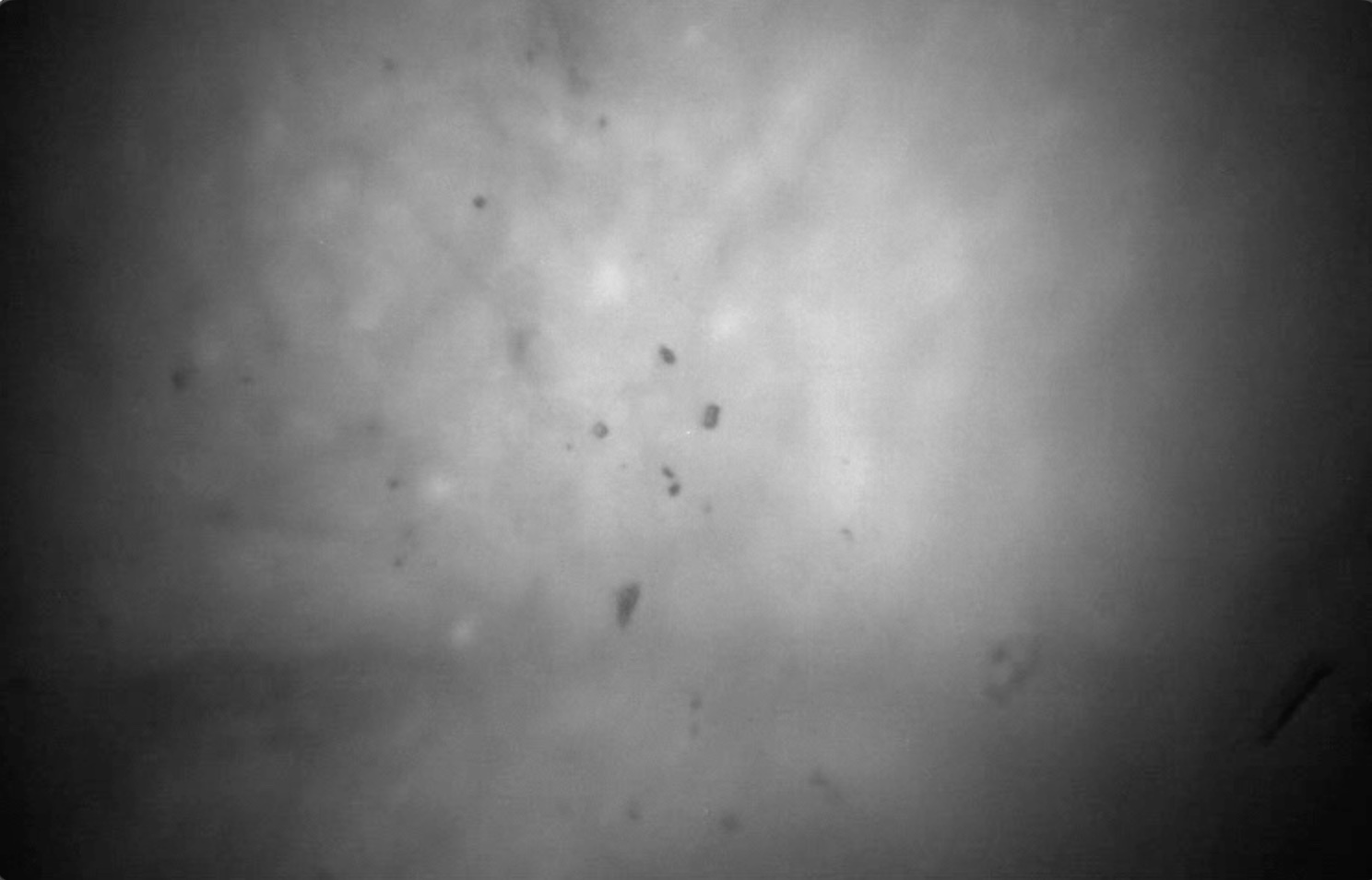} &
$\rightarrow$ \texttt{[[1,1,1],[1,1,0],[0,0,0]]} \\

Example Clip 3: &
\includegraphics[width=2.2cm,height=1.3cm]{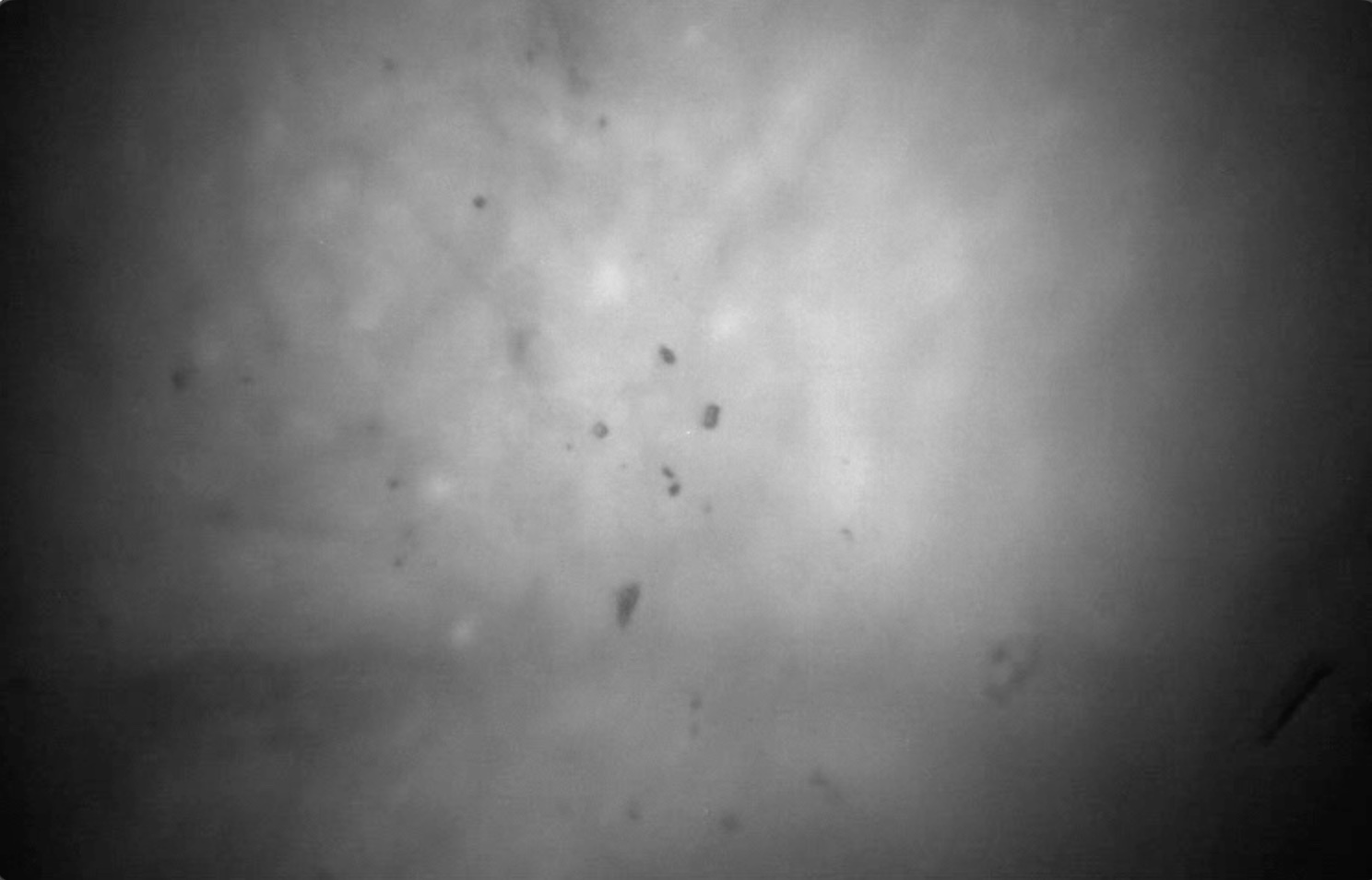} &
$\rightarrow$ \texttt{[[1,1,1],[1,1,1],[0,1,1]]} \\

Example Clip 4: &
\includegraphics[width=2.2cm,height=1.3cm]{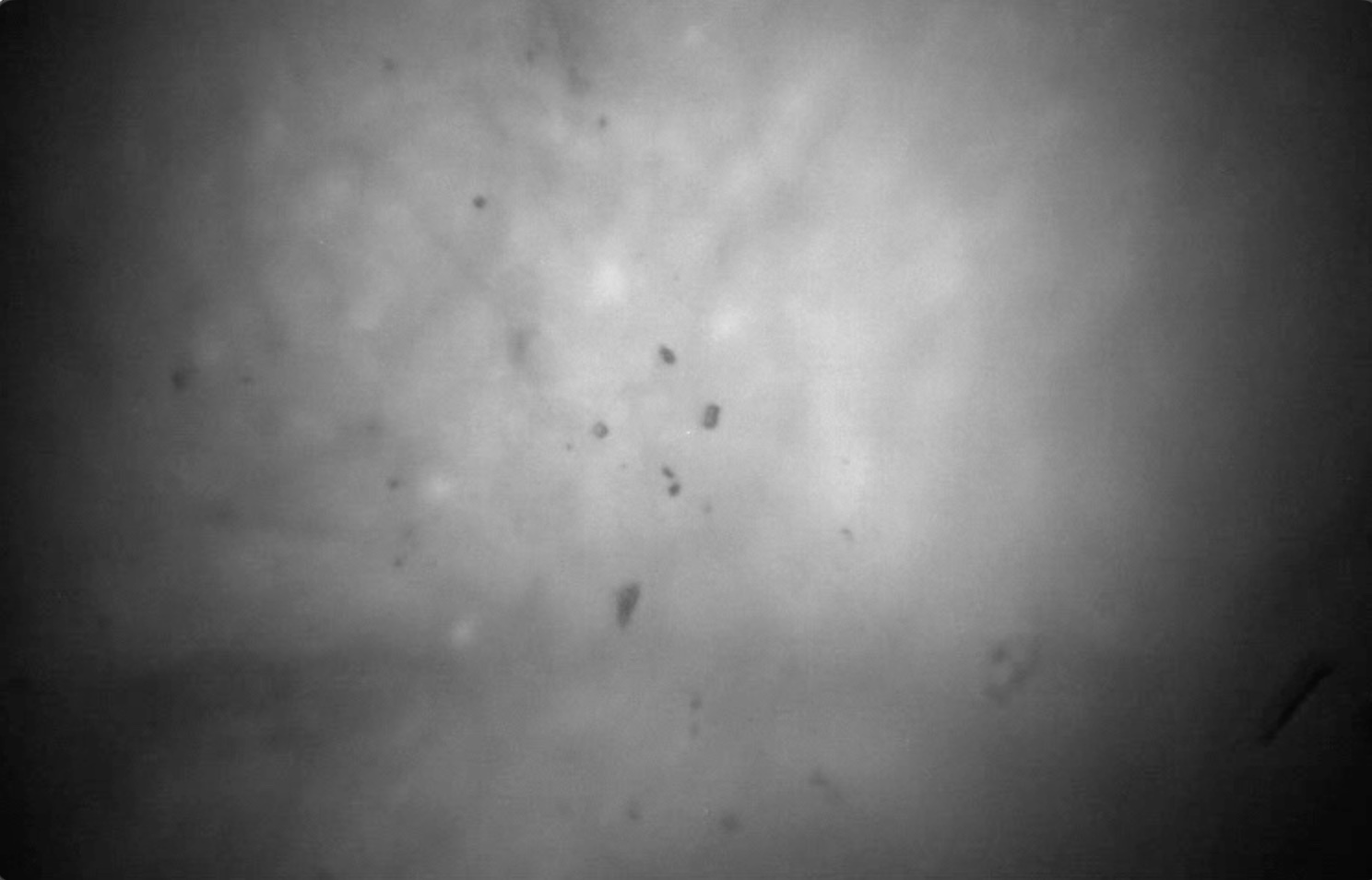} &
$\rightarrow$ \texttt{[[0,0,0],[0,0,0],[0,1,0]]} \\

Example Clip 5: &
\includegraphics[width=2.2cm,height=1.3cm]{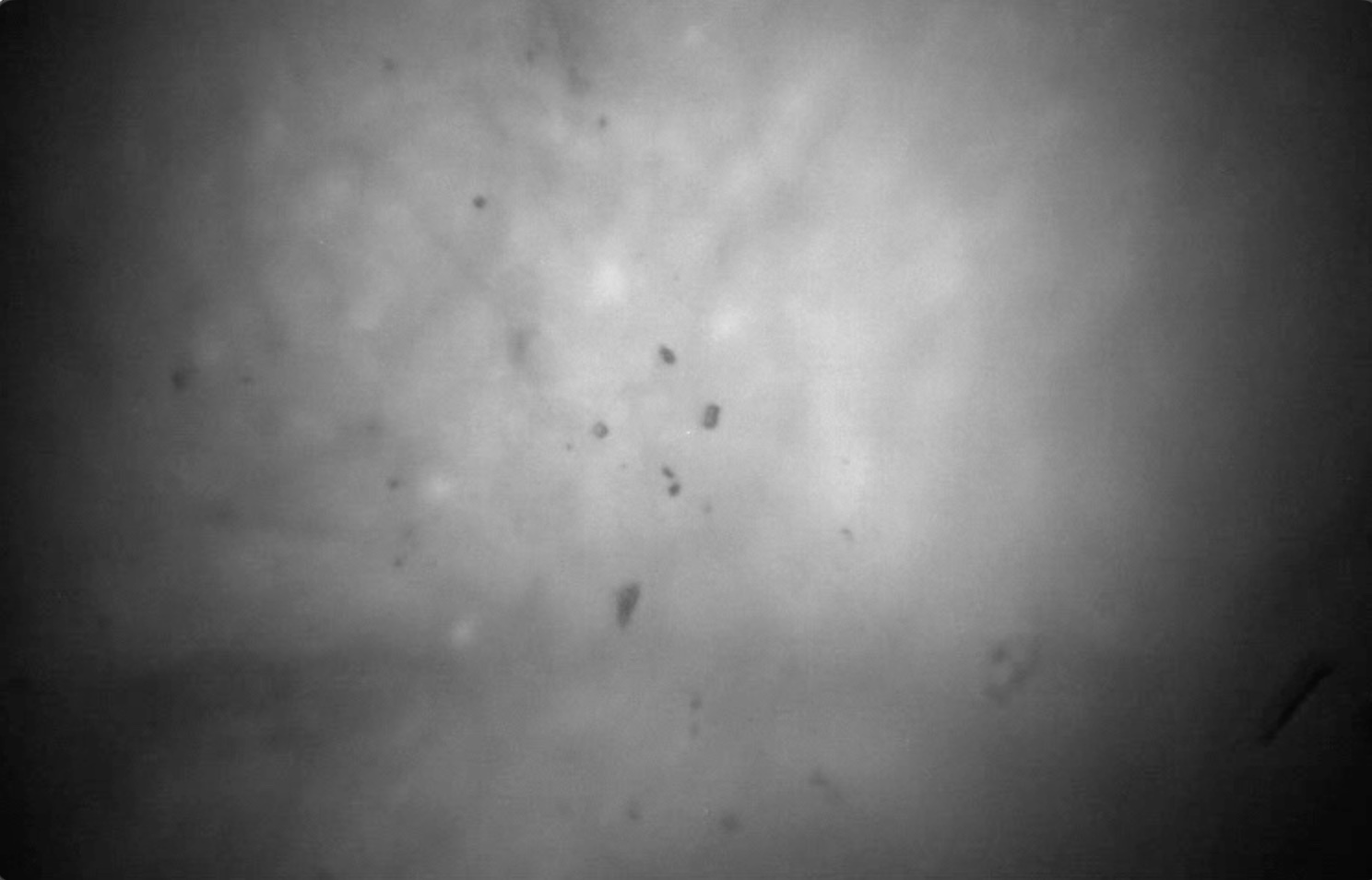} &
$\rightarrow$ \texttt{[[0,1,1],[0,1,0],[0,1,1]]} \\

Example Clip 6: &
\includegraphics[width=2.2cm,height=1.3cm]{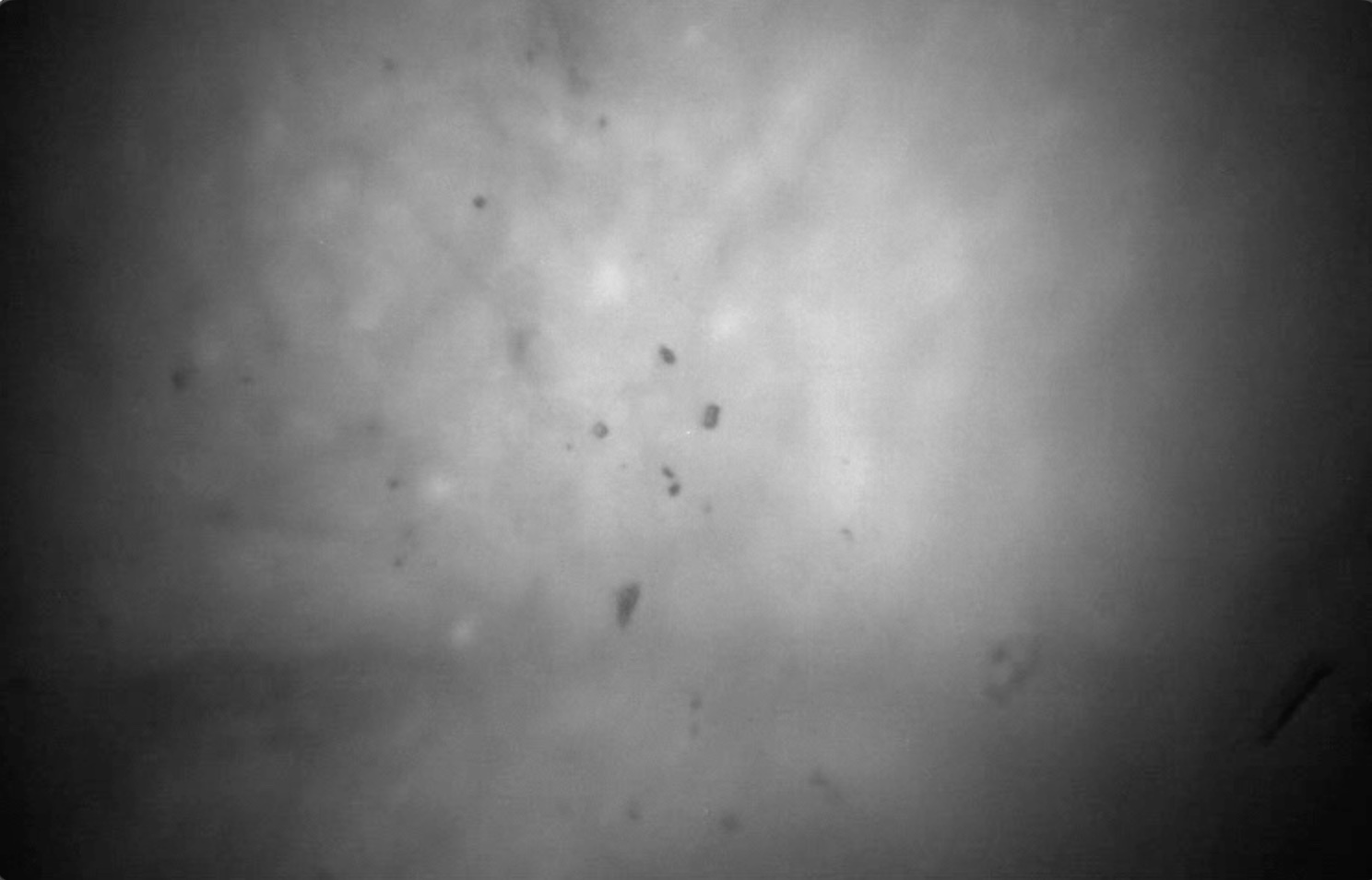} &
$\rightarrow$ \texttt{[[1,1,1],[0,1,0],[0,0,0]]} \\
\end{tabular}

\vspace{0.55cm}
\textbf{Target (Current Segment):}

\begin{tabular}{l c l}
Target Clip: &
\fbox{
\parbox[c][1.3cm][c]{2.2cm}{\centering \textbf{Input}}
} &
$\rightarrow$ \texttt{[ \ \ \ ]} \\
\end{tabular}

\end{minipage}
}
\caption{ICL prompt used for coarse calcium activity extraction. The VLM is given example calcium imaging clips paired with binary activity matrices over a spatial grid and predicts the activity matrix for a new input segment.}
\label{fig:calcium_icl_prompt}
\end{figure}

\section{Additional AI-driven Interpretations}

Here we include the full set of expert and VLM-generated interpretations for all latent components evaluated in the discovery experiment, along with their corresponding discovery scores. These are shown in Tables \ref{tab:full_discovery_results_part1}, \ref{tab:full_discovery_results_part2}, \ref{tab:full_discovery_results_part3}, \ref{tab:full_discovery_results_part4}, and \ref{tab:full_discovery_results_part5}. We also include the full expert and VLM interpretation from our full pipeline components as shown in \ref{tab:full_pipeline_discovery_appendix}.

\begin{table*}[!ht]
\centering
\caption{Full comparison of expert-written interpretations and VLM-generated hypotheses for latent components (Part 1 of 5).}
\label{tab:full_discovery_results_part1}
\small
\begin{tabular}{p{1.1cm} >{\raggedright\arraybackslash}p{6.2cm} >{\raggedright\arraybackslash}p{6.2cm} c c}
\toprule
\textbf{Comp.} & \textbf{Expert Interpretation} & \textbf{VLM Interpretation} & \textbf{Expert} & \textbf{VLM} \\
 &  &  & \textbf{Score} & \textbf{Score} \\
\midrule

1 & 
Looking at the third factor (across-trial factor), previous research supports the conclusion that a specific neuronal population will be engaged in all environments that an animal is exposed to if that animal does not associate any valence to the environments.  
Discovery Score = 5; this latent variable is supported by previous research &
Looking at the factor (across-trial factor), there is a clear pattern of neural engagement that occurs in three distinct, non-overlapping time windows, each marked by a burst of activity in a subpopulation of neurons. These bursts appear to be temporally separated and occur at approximately equal intervals across the trials, suggesting a periodic or phase-locked response to specific events or transitions within the behavioral paradigm. The activity is sparse and transient, with minimal engagement outside these windows.

This pattern is consistent with findings in the prelimbic cortex during learning tasks, where neural populations exhibit phasic, event-locked activity corresponding to discrete stages of learning, such as cue presentation, decision-making, or reward expectation. Specifically, studies have shown that prelimbic neurons can encode trial-specific events or transitions between behavioral states, often with sparse, transient bursts that are temporally aligned with task-relevant cues or outcomes. The three distinct peaks may reflect engagement during three distinct phases of learning or three separate trials within a session, where the animal is processing or updating its internal representation of the environment.

Discovery Score = 4; this latent variable is supported by previous literature, which demonstrates phasic, event-locked neural activity in the prelimbic cortex during learning, though the exact triphasic structure observed here is not explicitly reported in the literature. &
5 & 4 \\

\addlinespace
\midrule
\addlinespace

2 & 
Looking at the third factor (across-trial factor), previous research supports the existence of a subpopulation of neurons that are highly co-engaged within an environment that is very familiar to the animal and which the animal can recognize as its homecage.  
Discovery Score = 5; this latent variable is supported by previous research &
Looking at the factor (across-trial factor), there is a clear and progressive increase in neural engagement over the course of the behavioral sessions, with minimal activity at the beginning and a sharp rise toward the end, particularly in the final few trials. This suggests that a specific subpopulation of neurons becomes increasingly active as learning progresses, potentially reflecting the consolidation of a learned association or the emergence of a behavioral strategy. 

This pattern is consistent with findings in the prelimbic cortex during fear conditioning and extinction paradigms, where neural activity in this region is known to increase with learning and is associated with the formation of context-outcome associations. Specifically, studies have shown that prelimbic neurons exhibit ramping activity during learning, with increased firing rates as animals acquire task-relevant information, particularly during the transition from early to late stages of training. This ramping pattern is thought to reflect the encoding of predictive value or the updating of internal models.

Discovery Score = 4; this latent variable is supported by previous literature, particularly in the context of learning-related ramping activity in the prelimbic cortex, though direct evidence for this exact pattern in the same behavioral paradigm is not fully established. &
5 & 4 \\

\bottomrule
\end{tabular}
\end{table*}

\begin{table*}[!ht]
\centering
\caption{Full comparison of expert-written interpretations and VLM-generated hypotheses for latent components (Part 2 of 5).}
\label{tab:full_discovery_results_part2}
\small
\begin{tabular}{p{1.1cm} >{\raggedright\arraybackslash}p{6.2cm} >{\raggedright\arraybackslash}p{6.2cm} c c}
\toprule
\textbf{Comp.} & \textbf{Expert Interpretation} & \textbf{VLM Interpretation} & \textbf{Expert} & \textbf{VLM} \\
 &  &  & \textbf{Score} & \textbf{Score} \\
\midrule

3 & 
Looking at the third factor (across-trial factor), some research supports the notion that there is a reorganization of network activity, where a subpopulation of neurons might become highly active when the animal is in a novel, ambiguous environment that might signal safety or it might signal threat. However, other conclusions can be made that there should be a division of this population with one group of neurons co-engaged during exposure to the threatening environment, and a different subpopulation of neurons that become co-engaged when the animal is exposed to the safe environment, or the environment that does not pose a threat to the animal.  
Discovery Score = 3; this latent variable is supported by previous literature; however, there is also research that concludes the opposite as to what this component displays. &
Looking at the factor (across-trial factor), there is a distinct pattern of neural engagement that emerges during the middle portion of the behavioral session, with two prominent peaks of activity occurring in the second and third blocks of trials, followed by a sharp decline in activity in the final block. This suggests a transient population response that is not sustained across the entire session and may reflect a phase of active learning or consolidation. 

In the provided neuroscience literature, the prelimbic cortex is known to exhibit dynamic population coding during learning, particularly during the acquisition phase of associative tasks, where neural activity often increases as animals form cue-outcome associations. However, the specific pattern of two distinct peaks followed by a decline is not directly reported in the literature for this brain region or task. While some studies show increased activity during learning, the precise biphasic structure and subsequent suppression are not well-documented.

Scientific interpretation: This factor may represent a transient population response associated with the initial encoding of cue-outcome associations, followed by a reduction in activity as the task becomes more automated or as the animal transitions to a different cognitive state (e.g., consolidation or disengagement). The decline in the final block may reflect a shift in neural strategy or a reduction in attentional demand.

Discovery Score = 3; this latent variable is partially supported by literature, but the specific biphasic pattern is not directly reported. &
3 & 3 \\

\addlinespace
\midrule
\addlinespace

4 & 
Looking at the third factor (across-trial factor), there is high engagement of a group of neurons during the first day of the behavior paradigm within all 3 environments. There is currently no literature that supports the idea that there is a subpopulation of neurons that responds specifically to the first exposure of the 3 different environments and that this population would not respond to these 3 environments similarly 24 hours later. No associative learning is occurring between habituation exposure 1, 2, or 3 with the three exposures to all 3 environments separated by 24 hours.  
Discovery Score = 1; this latent variable is unsupported by current literature &
Looking at the factor (across-trial factor), the pattern shows two distinct peaks of neural engagement occurring at specific trials, with minimal activity in between, suggesting a sparse, event-locked response pattern. This may reflect a population of neurons that are selectively activated during discrete, salient events or transitions within the behavioral task, such as cue presentation or reward delivery. 

This pattern is consistent with findings in the prelimbic cortex during associative learning tasks, where specific neuronal ensembles are recruited during critical moments of learning, such as cue-outcome associations or decision points. The sparse, phasic activation aligns with known roles of prelimbic neurons in encoding task-relevant events rather than sustained environmental representation. 

Discovery Score = 4; this latent variable is supported by previous literature, particularly in the context of event-related encoding in the prelimbic cortex during learning, though direct evidence for this exact sparsity and timing pattern is not fully established. &
1 & 4 \\

\bottomrule
\end{tabular}
\end{table*}

\begin{table*}[!ht]
\centering
\caption{Full comparison of expert-written interpretations and VLM-generated hypotheses for latent components (Part 3 of 5).}
\label{tab:full_discovery_results_part3}
\small
\begin{tabular}{p{1.1cm} >{\raggedright\arraybackslash}p{6.2cm} >{\raggedright\arraybackslash}p{6.2cm} c c}
\toprule
\textbf{Comp.} & \textbf{Expert Interpretation} & \textbf{VLM Interpretation} & \textbf{Expert} & \textbf{VLM} \\
 &  &  & \textbf{Score} & \textbf{Score} \\
\midrule

5 & 
Looking at the third factor (across-trial factor), there is no discernable pattern of the engagement for the subpopulation of neurons. There seems to be no recognizable correlation to the exposure to the different environments or the time period across the behavioral paradigm where precise learning of cue associations occurs.  
Discovery Score = 1; this latent variable is unsupported by current literature &
Looking at the factor (across-trial factor), the pattern shows sparse and sporadic neural engagement across trials, with no consistent activation in any particular trial or group of trials. The majority of the data points remain near the baseline ($y \approx 0$), with only a few isolated peaks occurring at random trial positions. This suggests a lack of sustained or structured population activity, possibly reflecting noise, low signal-to-noise ratio, or a non-task-related neural state.

In the provided neuroscience literature, such sparse and unstructured activity patterns in the prelimbic cortex are not typically associated with learning or task-relevant processing. Instead, the literature emphasizes the emergence of structured, environment-specific, or reward-predictive population dynamics during learning. While some studies report transient or sparse activity during early learning phases, the complete absence of sustained or clustered engagement across trials is not a well-supported or commonly observed phenomenon in the context of associative learning paradigms.

Scientific interpretation: This factor may represent a non-informative or noise component, or perhaps a population that is not engaged in the task at all. It does not reflect a coherent neural representation of environment, valence, or learning state, and thus may not correspond to a meaningful latent variable in the context of the behavioral task.

Discovery Score = 1; this latent variable is unsupported by current literature. &
1 & 1 \\

\bottomrule
\end{tabular}
\end{table*}

\begin{table*}[!ht]
\centering
\caption{Full comparison of expert-written interpretations and VLM-generated hypotheses for latent components (Part 4 of 5).}
\label{tab:full_discovery_results_part4}
\small
\begin{tabular}{p{1.1cm} >{\raggedright\arraybackslash}p{6.2cm} >{\raggedright\arraybackslash}p{6.2cm} c c}
\toprule
\textbf{Comp.} & \textbf{Expert Interpretation} & \textbf{VLM Interpretation} & \textbf{Expert} & \textbf{VLM} \\
 &  &  & \textbf{Score} & \textbf{Score} \\
\midrule

6 &
High neuron coactivity during the threatening and familiar environments. Unexpected result, but might be due to (1) the mouse knows the familiar environment poses no threat, and (2) the threatening environment has a strong negative valence associated with it — where the animal is afraid (demonstrating high freezing behavior) throughout learning to discriminate the environments. &
Looking at the factor (across-trial factor), High neuron co-activity during early learning phase and sustained activity throughout the learning process, with peaks occurring in both safe and threatening environments, suggesting a general engagement during environmental exploration and initial learning.  
Discovery Score = 2; this latent variable is not directly supported by previous research, but some studies report broad prelimbic cortex activation during early learning stages. &
1 & 2 \\

\addlinespace
\midrule
\addlinespace

7 &
High neuron co-activity during habituation phase (when environments hold no meaning or valence to the mouse) and near the end of learning when the animal knows that the safe environment and threatening environments signal safety and threat, respectively. &
Looking at the factor (across-trial factor), High neuron co-activity during the early phase of learning, particularly before the first transition (around trial 10), with sustained but lower activity during the later phase of learning.  
Discovery Score = 3; this latent variable is supported by previous literature; however, there is also research that concludes the opposite as to what this component displays. &
2 & 3 \\

\addlinespace
\midrule
\addlinespace

8 &
Safe environment responding component, with some familiar environment response towards end of training when the mouse learns the safe environment poses no threat. &
Looking at the factor (across-trial factor), High neuron co-activity during the initial phase of learning (before environmental valence is established) and during the final phase of learning (when the animal has learned to discriminate environments), with minimal activity during the middle phase.  
Discovery Score = 3; this latent variable is supported by previous literature; however, there is also research that concludes the opposite as to what this component displays. &
4 & 3 \\

\addlinespace
\midrule
\addlinespace

9 &
Threating environment responding component, with high neuron co-activity during threatening environment exposure; especially near the end of the behavior paradigm, where the mouse knows the threatening environment signals threat. &
Looking at the factor (across-trial factor), High neuron co-activity during the threatening environment, with a peak occurring near the end of learning, suggesting increased engagement as the animal consolidates threat-related memory.  
Discovery Score = 4; this latent variable is supported by previous literature. &
4 & 4 \\

\bottomrule
\end{tabular}
\end{table*}

\begin{table*}[!ht]
\centering
\caption{Full comparison of expert-written interpretations and VLM-generated hypotheses for latent components (Part 5 of 5).}
\label{tab:full_discovery_results_part5}
\small
\begin{tabular}{p{1.1cm} >{\raggedright\arraybackslash}p{6.2cm} >{\raggedright\arraybackslash}p{6.2cm} c c}
\toprule
\textbf{Comp.} & \textbf{Expert Interpretation} & \textbf{VLM Interpretation} & \textbf{Expert} & \textbf{VLM} \\
 &  &  & \textbf{Score} & \textbf{Score} \\
\midrule

10 &
High neuron co-activity during familiar environment with some activity in the safe and threatening environments towards the end of learning (suggesting mouse is recognizing the environments?). &
Looking at the factor (across-trial factor), High neuron co-activity during the initial phase of learning and again during the later phase of learning, with minimal activity during the middle phase. This suggests a biphasic engagement pattern, possibly reflecting early encoding and late consolidation or reactivation processes.  
Discovery Score = 3; this latent variable is supported by previous literature; however, the specific biphasic pattern is not consistently reported and may reflect task-specific dynamics not fully captured in prior studies. &
3 & 3 \\

\addlinespace
\midrule
\addlinespace

11 &
Safe environment responding component, with high neuron co-activity during safe environment exposure, and also high activity during familiar environment exposure. &
Looking at the factor (across-trial factor), High neuron co-activity during the initial phase of learning and again near the end of learning, with minimal activity during the middle phase. This suggests a pattern of engagement during early encoding and late consolidation or decision-making stages.  
Discovery Score = 3; this latent variable is supported by previous literature; however, the specific timing and dual-peak structure are not consistently reported across studies. &
3 & 3 \\

\addlinespace
\midrule
\addlinespace

12 &
Unspecific activity, as high neuron co-activity is being detected across all 3 environment exposures and across learning epochs. &
Looking at the factor (across-trial factor), High neuron co-activity during the initial phase of learning (early trials) and again during the later phase of learning (after the second transition), suggesting a pattern of engagement during both early exploration and late consolidation or re-evaluation.  
Discovery Score = 2; this latent variable is not really supported by previous research. &
1 & 2 \\

\bottomrule
\end{tabular}
\end{table*}

\begin{table*}[!ht]
\centering
\caption{Full comparison of expert-written interpretations and VLM-generated hypotheses for components obtained from the end-to-end pipeline on two previously unlabeled subjects.}
\label{tab:full_pipeline_discovery_appendix}
\small
\begin{tabular}{p{1.1cm} >{\raggedright\arraybackslash}p{6.2cm} >{\raggedright\arraybackslash}p{6.2cm} c c}
\toprule
\textbf{Subject} & \textbf{Expert Interpretation} & \textbf{VLM Interpretation} & \textbf{Expert} & \textbf{VLM} \\
 &  &  & \textbf{Score} & \textbf{Score} \\
\midrule

1 &
Time 1 has high variability during the trial showing unspecific neuron activity for the majority of the environment exposure. However, a noticeably sharp increase towards the end of the trial stands out. This may indicate recognition of the environment. Alternatively, since this is the threatening environment, and the mouse did not receive the footshock it expected, this high activity may be due to a reward-type stimulus caused by the lack of threat. &
Looking at the factor (across-trial factor), Highly variable and noisy neural engagement with no clear temporal structure or alignment to behavioral epochs, suggesting either non-specific activity or potential artifact. Discovery Score = 1; this latent variable is unsupported by current literature. &
1 & 1 \\

\addlinespace
\midrule
\addlinespace

2 &
Time 0 has select time periods during the trial where neurons are highly co-engaged. This may be during different behavior output, such as the mouse engaging in freezing or grooming behavior. Although the values here cannot indicate valence of the environment, this component may be providing latent information about behavior planning and/or engagement during the trial. &
Looking at the factor (across-trial factor), High and variable neuron co-activity across the session, with multiple sharp peaks distributed throughout early, middle, and late phases, suggesting a non-specific or exploratory neural response pattern that does not align with distinct behavioral epochs or environmental valence. Discovery Score = 2; this latent variable is not well-supported by previous research, as most studies report more structured, environment- or learning-phase-specific dynamics in the prelimbic cortex. &
3 & 2 \\

\bottomrule
\end{tabular}
\end{table*}

\section{Behavioral Environments}

Here, we provide visual examples of the behavioral environments used during the fear conditioning and discrimination paradigm described. These environments were designed to differ in visual, tactile, and contextual cues while maintaining comparable spatial layouts, enabling mice to learn to discriminate between safe, threatening, and familiar contexts. These enviornments can be seen in Figure \ref{fig:appendix_environments}.

\begin{figure}[!ht]
    \centering
    \includegraphics[width=0.75\linewidth]{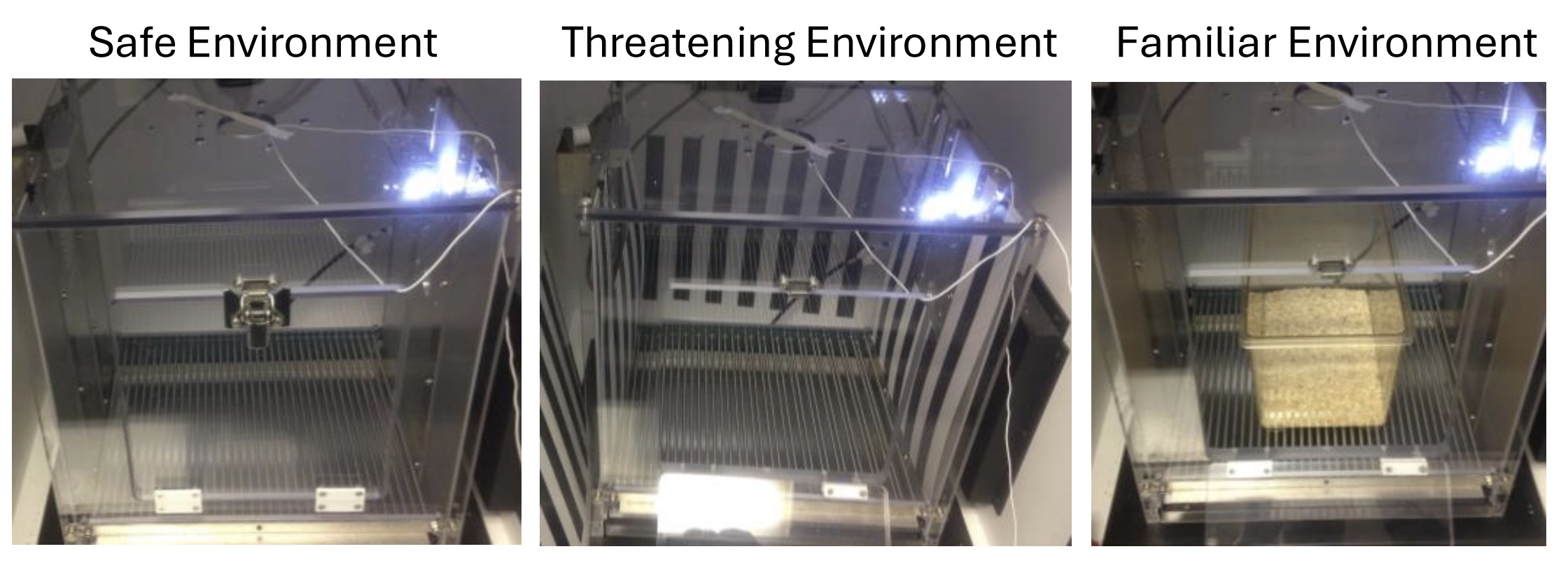}
    \caption{
    Behavioral environments used in the fear conditioning and discrimination task. 
    From left to right: \textbf{Safe environment (CS$-$)} which signals safety, 
    \textbf{Threatening environment (CS$+$)} paired with mild foot shocks during conditioning, 
    and a \textbf{Familiar home-cage analog (NS)} environment.
    }
    \label{fig:appendix_environments}
\end{figure}

\section{Tensor decomposition}
We provide tensor decomposition results on all subjects in \Cref{appendix:tab:tensor_methods}.
\begin{table}[h!]
\centering
\caption{Test RMSE of tensor completion. Note that C and NC indicate if tensors are coupled or not.
}
\label{appendix:tab:tensor_methods}
{\small
\begin{tabular}{l l cccc}
\toprule
& & \multicolumn{2}{c}{CPD}  & \multicolumn{2}{c}{NeAT}  \\
\cmidrule(lr){3-4} \cmidrule(lr){5-6} 
\multicolumn{2}{c}{\textbf{Subjects}}
& \multicolumn{1}{c}{NC} & \multicolumn{1}{c}{C}  
& \multicolumn{1}{c}{NC} & \multicolumn{1}{c}{C}  \\
\midrule
\multirow{7}{*}{\textbf{Control}}
     & 1 & 0.1457 & 0.2077 & \textbf{0.1099} & 0.1113 \\
     & 2 & 0.1255 & 0.1483 & 0.0937 & \textbf{0.0919} \\
    & 3 & 0.1231 & 0.1280 & \textbf{0.1002} & 0.1010 \\
    & 4 & 0.0928 & 0.1071 & \textbf{0.0724} & 0.0729 \\
    & 5 & 0.1067 & 0.1085 & \textbf{0.0781} & 0.0792 \\
    & 6 & 0.0980 & 0.1543 & \textbf{0.0733} & 0.0747 \\
    & 7 & 0.1443 & 0.1675 & \textbf{0.0944} & 0.0969 \\
\bottomrule
\end{tabular}
}
\end{table}

\end{document}